\documentclass[preprint,12pt]{elsarticle}

\usepackage{amssymb}
\usepackage{mathtools}
\usepackage{tcolorbox}
\usepackage{graphicx}
\usepackage{comment}
\usepackage{subfigure}
\usepackage{mathtools}
\usepackage{multirow}
\usepackage{multicol}
\usepackage{tabularx}
\usepackage{tabularray}
\usepackage{lscape}
\usepackage{float}
\usepackage[bb=px]{mathalpha}
\usepackage{url}
\usepackage{svg}
\usepackage{algorithm2e}
\usepackage{amsmath}
\usepackage{amssymb}
\usepackage{amsthm}
\newtheorem{definition}{Definition}

\journal{EURO Journal on Decision Processes}

\begin{document}

\begin{frontmatter}



\title{Leveraging Taxonomy Similarity for Next Activity Prediction in Patient Treatment}


\author[dfki]{Martin Kuhn\corref{cor1}}
\author[dfki,uni]{Joscha Grüger}
\author[dfki]{Tobias Geyer}
\author[dfki,uni]{Ralph Bergmann}

\affiliation[dfki]{organization={German Reseach Centre for Artificial Intelligence (DFKI), SDS Branch Trier},
            addressline={Behringstraße 21}, 
            city={Trier},
            postcode={54296}, 
            state={Rhineland Palatinate},
            country={Germany}}

\affiliation[uni]{organization={University of Trier, Business Information Systems II},
            addressline={Behringstraße 21}, 
            city={Trier},
            postcode={54296}, 
            state={Rhineland Palatinate},
            country={Germany}}

\cortext[cor1]{}

\begin{abstract}
The rapid progress in modern medicine presents physicians with complex challenges when planning patient treatment. Techniques from the field of Predictive Process Monitoring, like Next-activity-prediction (NAP) can be used as a promising technique to support physicians in treatment planning, by proposing a possible next treatment step. Existing patient data, often in the form of electronic health records, can be analyzed to recommend the next suitable step in the treatment process. However, the use of patient data poses many challenges due to its knowledge-intensive character, high variability and scarcity of medical data. To overcome these challenges, this article examines the use of the knowledge encoded in taxonomies to improve and explain the prediction of the next activity. Thus, this study proposes the TS4NAP approach, which uses medical taxonomies (ICD-10-CM and ICD-10-PCS) in combination with graph matching to assess the similarities of medical codes to predict the next treatment step. The effectiveness of the proposed approach will be evaluated using event logs that are derived from the MIMIC-IV dataset. The results highlight the potential of using domain-specific knowledge held in taxonomies to improve the prediction of the next activity, and thus can improve treatment planning, organizational management and decision-making by making the predictions more explainable.
\end{abstract}

\begin{graphicalabstract}
\end{graphicalabstract}

\begin{highlights}
\item Proposed approach integrates medical taxonomies (ICD-10-CM and ICD-10-PCS) with graph matching to enhance next-activity prediction in treatment processes.
\item Demonstrated the potential of domain-specific knowledge from taxonomies to improve prediction accuracy and explainability in medical treatment planning using MIMIC-IV data.
\end{highlights}


\end{frontmatter}


\section{Introduction}
\label{sec:introduction}
The landscape of modern medicine is characterized by remarkable advancements in diagnostics, therapies, and personalized care. Physicians and medical staff are at the forefront of this transformative era, equipped with a wide range of tools and treatment options to address diseases and improve patient outcomes. However, as medical science continues to evolve, so does the intricate nature of treatment planning. In the midst of this complexity, physicians and medical staff face an arduous challenge: planning the next steps in a patient's treatment regimen. This task is not only hindered by the intricate web of disease progression, but also by the significant effort required to navigate a vast sea of potential treatment options \cite{Mertens2019}. When analyzing healthcare processes, often patient pathways are examined retrospectively, after they have been concluded. Using this approach, significant opportunities presented by more proactive, forward-looking methodologies can be overlooked. These methodologies aim to assess patient pathways as they occur, offering predictions and insights into their future progression \cite{Mans.2015}. This is generally addressed in the field of Predictive Business Process Monitoring. This field is concerned with predicting aspects such as the next activity, a sequence of activities, the cycle time or the outcome of an ongoing process \cite{Francescomarino.2018}. In clinical settings, next-step predictions also have operational value. Knowing what is likely to occur next supports short-term capacity planning, staff assignment, and coordination across units. Recognizing these challenges faced by physicians, medical staff, researchers, and healthcare professionals have turned to innovative techniques to assist in making informed decisions \cite{Cartolovni2022}. One such technique that holds significant promise is next activity prediction (NAP), a method that leverages existing patient data to recommend the most appropriate next treatment step. Because predicted steps map directly to resource needs (rooms, devices, consumables, and specialist time), these predictions can be translated into actionable plans for managers and care coordinators. Since the medical domain is knowledge intensive \cite{Mertens2019}, the utilization of medical data by technological solutions presents challenges,  as it can lead to inaccurate conclusions and recommendations. This is aggravated by the fact that a lot of medical data is documented in poor quality \cite{Nguyen2022}, e.g., due to unstructured and heterogeneous documentation. However, there is also uniformly structured and coded data across different facilities and documentation systems. This includes the documentation of diagnoses (e.g., ICD-10-CM) and procedures (e.g., ICD-10-PCS), with the codes organized into taxonomies. These codes present knowledge, offering insights into entities and their interrelationships. By adopting the knowledge contained in taxonomies, decision support methods, such as NAP, can be semantically enriched. This can help to ensure that the proposed next treatment steps provide a medically sound and accurate basis for decision-making.

Therefore, this paper proposes an approach that utilizes the knowledge encoded in taxonomies to predict next activities in treatment processes. Leveraging patterns from previous patient cases to predict treatment decisions accurately and efficiently by not relying on black box approaches. Thus, the similarity-based TS4NAP (Taxonomic Similarity For Next Activity Prediction) approach is proposed. TS4NAP uses taxonomic similarities of activities, to compute the next possible activities and thus can support operational tasks like bed management, and ensuring the timely availability of equipment and staff. Thus, next-activity predictions can aid care-pathway coordination and short-horizon operations while complementing clinical judgment and governance processes. In parallel, the same predictions could also be used by physicians to make informed decisions about the next treatment steps. The approach is evaluated by utilizing the MIMIC-IV dataset, which is an expansive relational database that encompasses actual hospital stays of patients admitted to a medical center located in Boston, MA, USA \cite{Johnson2023}. The approach is a promising attempt to address medicine-specific characteristics and challenges for process-oriented environments presented in Munoz-Gama et al. \cite{Munoz-Gama2022}. On the one hand, the substantial variability (D1) is addressed due to the data used, and on the other hand, a white-box approach (D8) is applied to ensure the analysis of healthcare processes through transparency and understandability. Furthermore, the proposed approach is evaluated using real data (C4).

The remainder of the paper is organized as follows. Section \ref{sec:background} provides background information on event logs, taxonomies, similarity measures, information content, bipartite graph matching and related work. Section \ref{sec:methodology} describes the research method used to leverage the knowledge contained in taxonomies for predicting the next activity. Section \ref{sec:case_study} presents the case study where the MIMIC-IV event log is introduced, the  TS4NAP approach is evaluated, and the corresponding results are presented. Afterward, the results are discussed in Section \ref{sec:discussion} and possible directions for future work are given. Section \ref{sec:conclusion} concludes the publication.

\section{Background and Related Work}
\label{sec:background}

\subsection{Event Log and Taxonomies}
\label{sec:event_log}
Event logs can be viewed as multi-sets of cases. Each case consists of a sequence of events, i.e., the trace. Events are execution instances of activities. Here, the execution of an activity can be represented by multiple events. Beyond the scope of control-flow analysis, event logs can also incorporate attributes to symbolize alternative perspectives, including the data perspective. In the context of event logs, taxonomies serve as classification systems that help to map activities to the concepts defined by the taxonomy. Thus, it is possible to include multiple taxonomies for one event log, where each different activity could be represented by a different taxonomy. These taxonomies can then be used to enrich the activities with coded domain knowledge, which then could be used for various process mining related analysis. In the following, event logs, traces, events and taxonomies are defined. \cite{dataScienceinAction}
\begin{definition}(Universes) For this paper, the following universes are defined \cite{dataScienceinAction}. 
\begin{itemize}
    \item The universe of all possible case identifiers $\mathcal{C}$.
    \item The universe of all possible event identifiers $\Sigma$.
    \item The universe of all possible activity identifiers $\mathcal{A}$.
    \item The universe of all possible attribute identifiers $\mathcal{AN}$.
    \item $\mathcal{T}$ defines the universe of taxonomies.
    \item $\mathcal{P}$ defines all the possible lists of categorical values.
\end{itemize} 
\end{definition}

\begin{definition}(Trace, Case) Each case $c\in\mathcal{C}$ has a mandatory attribute \textit{trace}, with $\hat{c}=\#_{trace}(c)\in\Sigma^*\setminus \{\langle\rangle\}$, where $\#_{n}(c)$ retrieves  the value for the attribute $n\in\mathcal{AN}$ for the case $c$. A trace is a finite sequence of events $\sigma\in\Sigma^*$ where each event occurs only once, i.e. $1 \leq i < j \leq | \sigma |: \sigma(i) \neq \sigma(j)$. \cite{dataScienceinAction}
\end{definition}
\begin{definition}(Event log, Events) An event log is a set of cases $\mathcal{L}\subseteq\mathcal{C}$, in the form that each event $e\in\Sigma$ is contained only once, in the event log. If an event log contains timestamps, these should be ordered in each trace. \cite{dataScienceinAction}
\end{definition}

\subsection{Medical Taxonomies}
\label{sec:medical_taxonomies}
A \textit{taxonomy} is a hierarchical classification system used to organize and categorize concepts or entities based on their shared characteristics or attributes. It provides a structured framework for organizing information, allowing for systematic classification and arrangement of various elements within a specific domain or field of study. Two taxonomies in the medical field are ICD-10-CM\footnote{\label{icd-10}\url{https://www.cms.gov/medicare/coding-billing/icd-10-codes}} and ICD-10-PCS\footref{icd-10}.

The International Classification of Diseases (ICD) system, developed by the World Health Organization (WHO), plays a pivotal role in documenting treatments and facilitating the payment process across many nations. It offers a unified framework for the reporting and surveillance of diseases and health conditions, significantly aiding in global health endeavors \cite{Hirsch2016Apr}. Furthermore, the ICD coding system is instrumental in healthcare reimbursement and resource allocation in numerous countries. Notably, nations such as the United States, Australia, Germany, and Canada have developed national extensions of ICD-9 or ICD-10 to cater to their specific healthcare requirements \cite{Cartagena2015Mar}.

\begin{figure}[htb]
\centering
  	{\includegraphics[width=.45\textwidth]{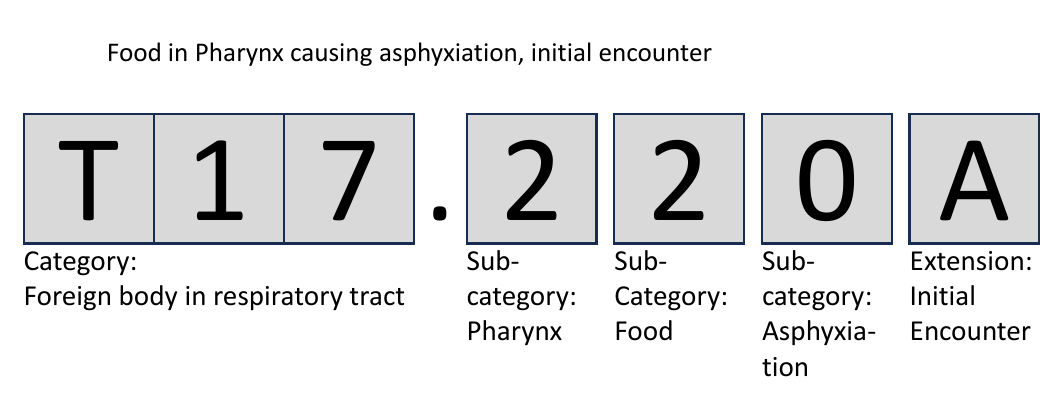}}
	\caption{ICD-10-CM structure with the example diagnosis of “Food in pharynx causing asphyxiation, initial encounter”\label{fig:icd10cm}}
\centering
\end{figure}

In the United States, the ICD-10 system is divided into two complementary subsets: ICD-10-CM (Clinical Modification), used for diagnostic coding, and ICD-10-PCS (Procedure Coding System), employed for coding inpatient hospital procedures. ICD-10-CM codes vary in length from four to seven characters, with the possibility of the sixth or seventh character being either alphabetic or numeric, enhancing the granularity of the diagnostic data \cite{Hirsch2016Apr}. Figure \ref{fig:icd10cm} shows the structure of the code along with an example. It can be seen that the first three characters are defining the category of the diagnosis. The subcategories define the diagnoses in more detail, and the last character defines the extension.

\begin{figure}[htb]
\centering
  	{\includegraphics[width=.45\textwidth]{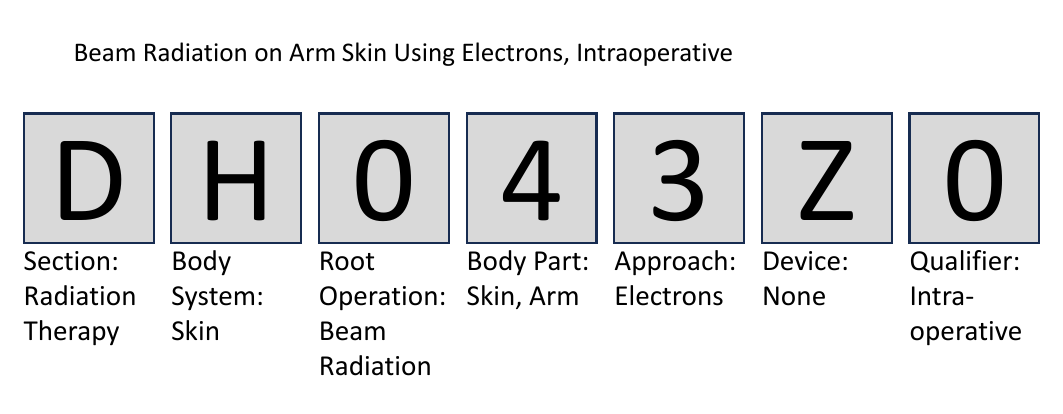}}
	\caption{ICD-10-PCS structure with example procedural code of “Beam Radiation on Arm Skin Using Electrons, Intraoperative" \label{fig:icd10pcs}}
\centering

\end{figure}

ICD-10-PCS constitutes a specialized coding system for procedural data, characterized by its alphanumeric, seven-character codes. Each character within an ICD-10-PCS code represents a specific aspect of the procedural information, such as the anatomical location or surgical approach, thereby encapsulating comprehensive details of the procedure. This structuring enables a multifaceted description of medical procedures, with the coding hierarchy beginning at a broad section level and narrowing down through various descriptors—ranging from the body system involved to the specific operation performed—culminating in a final qualifier that adds further specificity \cite{ICD10PCS, herandezsICD, FernandezICD}. Figure \ref{fig:icd10pcs} shows the structure of the ICD-10-PCS. 

These codes are specifically used for documenting inpatient procedures, playing a crucial role in the billing process for hospitals. They allow for detailed and precise recording of surgical operations and other medical procedures, ensuring accurate reimbursement and statistical analysis.

\subsection{Semantic Similarity in Taxonomies and Information Content}
\label{sec:Taxonomy_Background}
To measure or assess the degree of similarity or relatedness between different categories or concepts within a taxonomy, semantic similarity measures can be used. 
One possibility is based on approaching the concept of semantics in terms of \textit{information content} (IC). In the field of computational linguistics, the concept of IC plays a vital role in measuring the amount of embedded information within linguistic elements \cite{Pedersen2007}. Typically, concrete and specialized entities in a discourse tend to have higher IC compared to general and abstract ones. The IC can be used in similarity functions to compute the similarity between concepts contained in a taxonomy, according to the amount of information they share \cite{Sanchez2011}. In this case, the accurate estimation of the IC of concepts is crucially relied upon in estimating their similarity. Therefore, to compute meaningful values from a semantic perspective, all the explicit appearances of the concept and the appearances of concepts that are semantically subsumed by the concept (\textit{subsumer}) must be considered \cite{Sanchez2011}. For example, to estimate the IC of the concept 'surgery', all its explicit occurrences should be considered, along with the occurrences of all its specializations such as 'orthopedic surgery', 'cardiac surgery', and 'neurosurgery'.
In terms of taxonomies, semantic commonalities among concepts
based on the amount of information they share is represented by the \textit{least common subsumer} of both concepts \cite{Sanchez2011}. The LCS is the most specific taxonomical ancestor common to two concepts. Therefore, the higher the IC of the subsumer of both concepts, the greater the similarity between these two.

\subsection{Bipartite Graph Matching}
\label{sec:Graph_Matching}
Bipartite graph matching is a concept in graph theory with applications in various fields. A bipartite graph is a mathematical structure consisting of two disjoint sets of vertices, which can be used to explore and analyze the relationships between this set of vertices.  Matching within these graphs involves pairing vertices from one set with vertices in the other set according to certain criteria, often aiming to maximize or minimize the total weight of the matched pairs \cite{Asratian1998}. 

\begin{definition}(Bipartite Graph) 
\label{eqn:bipartitegraph}
A bipartite graph consists of two disjoint sets of vertices, \(U\) and \(V\), where every edge connects a vertex from \(U\) to a vertex in \(V\). Formally, a bipartite graph can be denoted as \(G = (U, V, E)\), where \(E\) represents the set of edges such that \(e = (u, v)\) for \(u \in U\) and \(v \in V\) and \(U \cap V = \emptyset\) holds. The set of all bipartite graphs $G$ will be denoted as $\mathcal{G}$.
\end{definition} 

A matching \(M\) in a bipartite graph is a set of edges such that no two edges share a common vertex. In other words, each vertex is incident to at most one edge of the matching. In weighted bipartite graphs, where a weight function assigns a real number (weight) to each edge, a maximum weight matching is a matching that maximizes the sum of the weights of the edges in the matching \cite{Asratian1998}.

\begin{definition}(Matching)
\label{eqn:matching}
A matching \(M\) in a bipartite graph \(G\) is a subset of \(E\) such that no two edges in \(M\) share a common vertex. This can formally be denoted as \(M \subseteq E\) \cite{Asratian1998}.
\end{definition}

\begin{definition}(Maximal Weight Matching in Bipartite Graphs)
\label{eqn:max_matching}
Each edge \(e \in E\) of a bipartite graph \(G = (U, V, E)\) has an associated weight \(w(e)\), which can represent the cost or benefit of including that edge in the matching. A weight function \(w: E \rightarrow \mathbb{R}\) assigns a weight to each edge. The set of all weight functions $w$ is defined as $\mathcal{W}$. A maximum weight matching is a matching \(M_{\text{opt}}\) that maximizes the sum of the weights of the edges in \(G\). Which can be denoted as: \(\sum_{e \in M_{\text{opt}}} w(e) \geq \sum_{e \in M} w(e), \text{ for all M} \subseteq E\) \cite{Asratian1998}. 
\end{definition} 

\begin{definition}
\label{eqn:maximum_weight_matching_function}
The maximum weight matching function $(mwm)$ is defined as $mwm : \mathcal{G} \times \mathcal{W} \rightarrow\mathbb{R}$. The function gets the bipartite graph $G$ and a weight function $w$ as input, and outputs the aggregated weight of the optimal mapping based on the function in Definition \ref{eqn:max_matching}.
\end{definition}

To find the maximum weight matching in a weighted bipartite graph, the Hungarian algorithm, or Kuhn-Munkres algorithm, could be utilized \cite{Kuhn1955, Munkres1957}. This algorithm transforms the problem into one of finding an assignment that minimizes or maximizes the total weight of the matching. The algorithm has an iterative improvement strategy. Starting with an initial feasible matching, it systematically adjusts the weights assigned to vertices to reveal new potential edges for matching, thus exploring the solution space efficiently. By identifying augmenting paths, it guarantees the optimization of the total weight of the matching. An augmenting path is a path that starts and ends at unmatched vertices, with edges alternating between matched and unmatched. This process continues until no further improvements can be made, ensuring an optimal solution is reached. The algorithm has a time complexity of $O(n^{4})$, for a problem of size $n$ \cite{Kuhn1955, Munkres1957}. There are also newer, more time efficient algorithms, which can be used to solve this problem \cite{Galil1983}.

\subsection{Related Work}
\label{sec:related_work}
Predictive Process Monitoring (PPM) focuses on forecasting the future evolution of running process instances, with next-activity prediction NAP as a central task alongside remaining-time and outcome prediction \cite{Mans.2015,Francescomarino.2018}. Most approaches follow a two-phase setup in which models are trained on historical event logs and then queried online on partial traces \cite{Ceravolo2024Dec}. Three families of techniques have emerged. First, Sequence-based methods, that treat a case history as an ordered string and learn next-step probabilities from observed subsequences, examples include Markovian and $n$-gram style estimators \cite{10.1109/iCCECE46942.2019.8941917} as well as early event-driven frameworks for online prediction \cite{Becker.2014,Polato.2018}. Second, Model-driven methods, discover an explicit control-flow model (e.g., Petri nets, BPMN, transition systems) and then use it for inference. For instance, stochastic Petri nets have been used to predict temporal properties \cite{10.1016/j.is.2015.04.004}, and process-model and attribute information have been exploited to estimate performance measures \cite{Verenich2019Jun}. While such approaches offer structure and interpretability, they are challenged by discovery or manual modeling in highly variable domains like healthcare \cite{Mannhardt.2017,grueger.2022}. Finally, deep learning has become dominant for NAP. Here, architectures like LSTM or RNN are utilized to model event sequences \cite{10.1007/978-3-031-08848-3_10,10.1007/978-3-319-59536-8_30} or encoder–decoder models predict multiple future steps \cite{10.1137/1.9781611975673.14}. Comparative studies report strong accuracy for deep models across public benchmarks \cite{Rama.2021,Tax.2020} and also show that these models are applicable in clinical settings \cite{Xu.2020,Ramirez-Alcocer.2023}.

Furthermore, context awareness emerges as a decisive driver of next-activity prediction performance. Context can be injected explicitly as additional features or more tightly coupled to the decoding procedure in deep learning models \cite{Heinrich.2021,10.1007/978-3-030-26619-6_19}. However, in healthcare, two constraints complicate the straightforward adoption of these models. First, pathways are highly variable and knowledge-intensive. Second, clinical decision support must be transparent. Clinicians are reluctant to act on predictions that are not clearly explainable \cite{Janiesch.2021,Shickel.2017,Munoz-Gama2022}. As a response, a complementary line of work adapts explainable AI (XAI) to PPM, providing post-hoc explanations or local feature attributions for predictions made by deep-learning models \cite{Bauer.2023,Weinzierl.2020,Mehdiyev.2021,Zilker.2023}. While such techniques can mitigate the black-box character of these models, they do not change the underlying black-box nature and may fall short of the accountability and traceability required in clinical settings \cite{Munoz-Gama2022}.

Interpretable, alternatives for NAP also exist. Model-based predictors grounded in discovered process structures preserve a direct link between behavior and inference \cite{10.1016/j.is.2015.04.004,Verenich2019Jun}. Similarity- and rule-based approaches provide another more explainable approach. For example, Maggi et al. retrieve cases by prefix similarity and trains on the retrieved subset to estimate goal-related outcomes, yielding explanations grounded in nearest examples \cite{Maggi2014}. In the clinical realm, case-based reasoning (CBR) aligns with practice by reusing the most similar prior patient trajectories in a transparent manner \cite{10.1016/j.artmed.2005.10.008}. Moreover, biomedical ontologies and taxonomies offer a technique to encode domain constraints and semantics, enabling knowledge-aware representations and matching \cite{10.1093/bib/bbaa199}. These methods directly address the interpretability concerns raised above, although they can struggle when confronted with profound variability, heterogeneous documentation, and noisy data \cite{Nguyen2022}.

Positioned within this domain, this work targets the intersection of context-awareness and interpretability for NAP in healthcare. Despite the breadth of prior research, two gaps remain. First, the rich semantics captured by medical taxonomies (e.g., hierarchical relations among diagnoses and procedures) are rarely exploited for next-step prediction. When ontologies or taxonomies appear, they often serve as auxiliary features or hard constraints rather than forming the core of the predictive mechanism \cite{10.1093/bib/bbaa199}. Second, few approaches jointly leverage (i) taxonomy-grounded similarity on both control-flow events and clinical context (e.g., multiple diagnoses) and (ii) an order-aware alignment that tolerates variability in event positions, all within a deterministic, explainable scoring scheme. This approach addresses these gaps by computing semantic similarity via information-content–based measures grounded in taxonomy structure and by aligning traces through maximum-weight bipartite matching to respect sequence order while accommodating clinical variability. The result is a ranked set of next-activity candidates accompanied by trace-level similarity scores that can be inspected, thereby directly supporting the transparency demanded by clinical stakeholders \cite{Munoz-Gama2022}. 

\section{TS4NAP: Taxonomy Similarity for next Activity Prediction}
\label{sec:methodology}
The TS4NAP (Taxonomy Similarity for next Activity Prediction) approach provides a framework for assessing similarity based on shared taxonomic characteristics and relationships for events, which are used for next activity prediction. The next activity is predicted by utilizing maximum graph matching for bipartite graphs to identify the most similar traces in an event log according to a query trace. Here, two traces are interpreted as bipartite graphs. The identified most similar traces are directly used for predicting a set of
possible next activities. 

\begin{figure}[htb]
\centering
  	{\includegraphics[width=.45\textwidth]{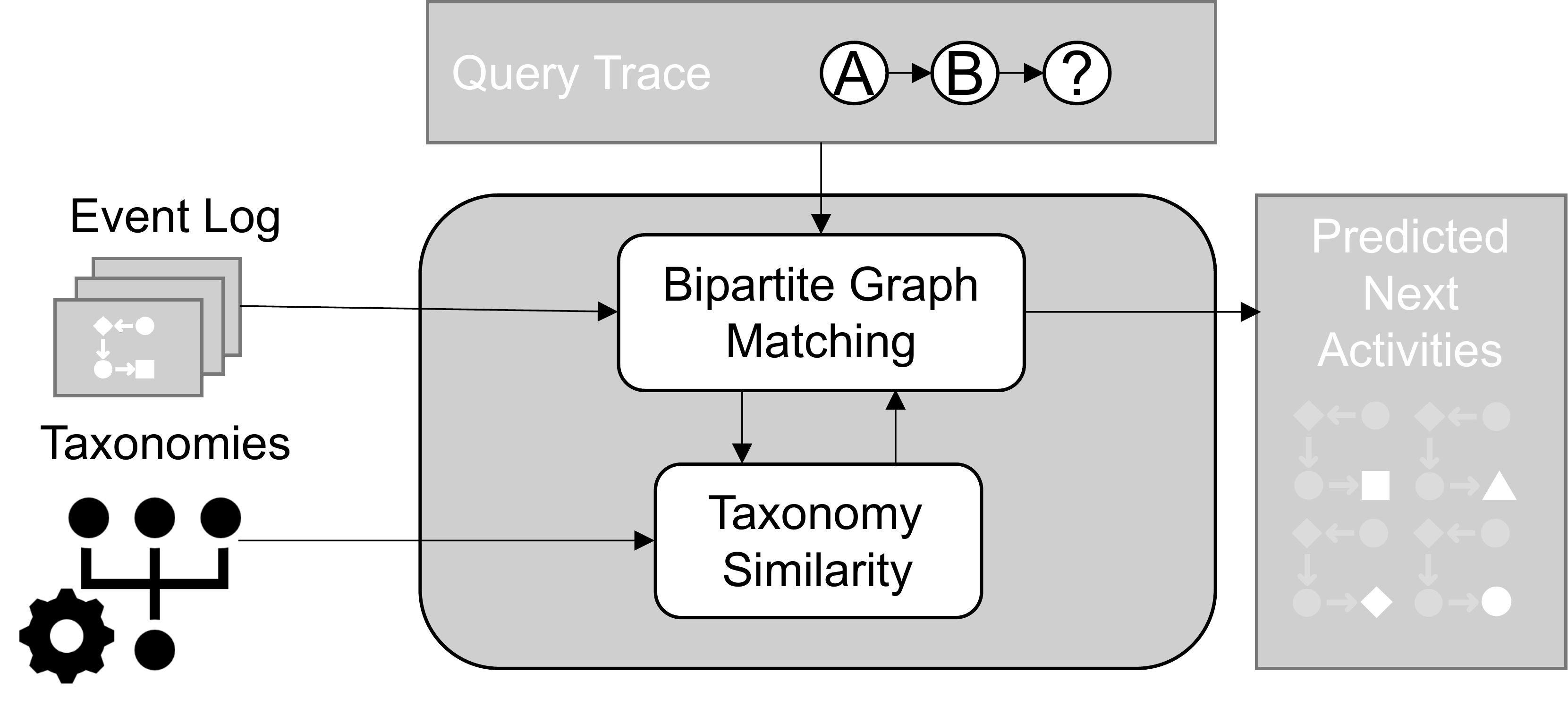}}
	\caption{The TS4NAP approach is based on calculating the semantic similarity between events using taxonomies. Thus, similar traces can be identified, which can then be used to propose a selection of the next activities. \label{fig:TS4NAP}}
\centering
\end{figure}

Figure \ref{fig:TS4NAP}, shows a visualization of the TS4NAP approach. The following discusses the concept of taxonomy-based semantic similarity, its application for calculating similarity in process data, and the TS4NAP approach, which leverages this method for predicting the next activity.

\subsection{Semantic similarity in Taxonomies}
\label{sec:Tax_sim}
In the TS4NAP approach, the similarity measure proposed by Sánchez et al. \cite{Sanchez2011} is utilized. This measure is an adaptation of a similarity measure proposed by Lin \cite{Lin1998}, which presents a universally applicable definition of similarity grounded in information theory. The adaptation by Sánchez incorporates the use of information content (IC) for improving the similarity measure, which was shown in their study \cite{Sanchez2011}. The IC measure proposed by Sánchez was specifically chosen because it can effectively distinguish concepts with similar counts of hyponyms or leaves but varying degrees of concreteness. By considering the number of subsumer associated with each concept, the proposed information content measure provides a more concrete, refined and accurate representation of the underlying semantic relationships within a concept hierarchy. Thus, it is less dependent on the taxonomic design than other methods like Resnik \cite{Resnik1995} and Seco et al. \cite{Seco2004}.
\begin{equation}
     IC(c)=-log\left(\frac{\frac{|leaves(c)|}{|subsumers(c)|}+1}{max\_leaves+1}\right)
    \label{eqn:sanchez_ic}
\end{equation}
The IC according to Sánchez is calculated by Equation \ref{eqn:sanchez_ic}. Here $leaves(c)$ describes the set of concepts, which are found at the end of the taxonomic tree under concept $c$. The complete set of taxonomic ancestors is denoted by $subsumer(c)$. In the case of multiple inheritance, all ancestors are considered. The fraction in the negative logarithm function is normalized by utilizing the total number of leaves in the taxonomy, which is expressed by $max\_leaves$. To normalize the fraction such that values in the range of $0$ to $1$ are produced, the value of $1$ is added to the numerator and denominator. \cite{Sanchez2011}

Equation \ref{eqn:Adapted_Lin} presents the similarity measure adapted by Sanchez, which extends Lin's similarity measure \cite{Lin1998} by incorporating the IC, shown in Equation \ref{eqn:sanchez_ic}. Equation \ref{eqn:Adapted_Lin} quantifies the relationship between the shared information of concepts and the total information required to fully describe them \cite{Sanchez2011}. This study refers to Equation \ref{eqn:Adapted_Lin} as Sánchez similarity, or in short $sim_{\textsubscript{Sánchez}}$. In the equation, the term $IC(LCS(c_{1}, c_{2}))$ represents the shared information between two concepts, denoted as $c_{1}$ and $c_{2}$. This shared information is quantified by utilizing the Least Common Subsumer LCS of the two concepts. 
\begin{equation}
       sim_{\textsubscript{Sánchez}}(c_{1}, c_{2}) = \frac{2 \cdot IC(LCS(c_{1}, c_{2}))}{IC(c_{1}) + IC(c_{2})}
       \label{eqn:Adapted_Lin}
\end{equation}
To achieve normalization, the information content IC of both compared concepts is summed up in the denominator. The denominator captures the total information required to describe both concepts adequately. In the numerator, the shared information $IC(LCS(c_{1}, c_{2}))$ is multiplied by two to normalize the measure. As a result, the adapted similarity measure yields values ranging between 0 and 1, indicating the degree of similarity between the concepts $c_{1}$ and $c_{2}$.

\subsection{Bipartite Graph Transformation}

The core idea of the approach is to compare two traces that exhibit variability in the sequence of events. To account for this variability in the similarity computation, maximum graph matching for bipartite graphs is performed. This necessitates transforming the traces to be compared into bipartite graphs. In addition to control flow data (i.e., the actual events), our study has encountered patient process data, which include lists of patient diagnosis that can be mapped to taxonomies. These list-based data, often ordered by priority, must be considered in the similarity calculation. Therefore, both control flow and list data are transformed into bipartite graphs.

\subsubsection*{Control Flow}  
To calculate the similarity between two traces $\sigma, \sigma' \in \mathcal{L}$, the traces are transformed into a bipartite graph \(G^{\sigma}_{\sigma'} = (U, V, E)\) (see Definition \ref{eqn:bipartitegraph}):
\begin{itemize}
    \item \( U = \{e_1, e_2, \dots, e_m\} \) represents the set of events from \( \sigma \),
    \item \( V = \{e_1', e_2', \dots, e_n'\} \) represents the set of events from \( \sigma' \),
    \item $E \subseteq U \times V$ is the set of edges such that
    \[
    E = \{ (e_i, e'_j) \mid e_i \in U, e'_j \in V \}.
    \]
\end{itemize}
This transformation is essential for computing the similarity between events, taking into account differences in event sequences.

\subsubsection*{Lists}  
To compute the similarity between two lists $\phi, \phi' \in \mathcal{P}$,the lists are similarly transformed into a bipartite graph \(G^{\phi}_{\phi'} = (U, V, E)\) (see Definition \ref{eqn:bipartitegraph}):
\begin{itemize}
    \item \( U = \{e_1, e_2, \dots, e_m\} \) represents the set of list elements from \( \phi \),
    \item \( V = \{e_1', e_2', \dots, e_n'\} \) represents the set of list elements from \( \phi' \),
    \item $E \subseteq U \times V$ is the set of edges such that
    \[
    E = \{ (e_i, e'_j) \mid e_i \in U, e'_j \in V \}.
    \]
\end{itemize}
This transformation is required to calculate the similarity between list elements, considering variations in the sequence of list items. Since the list data is always related to a trace $\sigma$ we use the shorthand $\sigma_{\phi}$ for accessing the list data $\phi$ of $\sigma$.

\subsection{Similarity Based on Maximum Weight Matching}
\label{sec:TS4NAP_new}
 Each node $e\in U\cup V$ is related to a concept of a taxonomy, represented by $e_\mathcal{T}$. The taxonomy-based similarity of two nodes $u,v\in U\cup V$ can then be lavaged by
\[
sim_{\textsubscript{Sánchez}}(u_\mathcal{T}, v_\mathcal{T})
\]
Since the most similar events between two traces or the most similar elements between two lists are not necessarily in the same position, but the order still holds significance, an order-based weight is assigned to each edge in the bipartite graph to reflect this importance.
\begin{itemize}
    \item $pos: {U \cup V} \rightarrow \mathbb{N}^{+} $ returning the temporal or priority positions of the events in the trace or the list elements in the list for the corresponding edge.
    \item $w_{order}(e) = 0.5^{|pos(u) - pos(v)|}$ is returning the order-based weight of the edge $e$ depending on the positions of $u$ and $v$.
    \end{itemize}
The taxonomic similarity and order-based weights can now be used to assign a taxonomic similarity and order-based weight to each edge in the bipartite graph, denoted by $w_{tax}: E \rightarrow \mathbb{R}$:
\[w_{tax}(e) = sim_{\textsubscript{Sánchez}}(u_\mathcal{T}, v_\mathcal{T})) * w_{order}(e)\]

Finally, we use the taxonomic weight function $w_{tax}$ and the bipartite graph $G^{\sigma}_{\sigma'}$ to map the nodes of the two given subsets $U, V$ of the bipartite graph onto each other using the maximum weight matching function $mwm$ (see Definition \ref{eqn:maximum_weight_matching_function}), and compute the weighted similarity of the control flow of traces by $sim_{cf}:\Sigma^*\times\Sigma^*\rightarrow\mathbb{R}$. For $\sigma,\sigma'\in\Sigma^*$ it holds:
    \[
    sim_{cf}(\sigma,\sigma') = \frac{1}{|\sigma|} mwm(G^{\sigma}_{\sigma'}, w_{tax})
    \]

In the same way, we use the taxonomic weight function $w_{tax}$ and the bipartite graph $G^{\phi}_{\phi'}$ to map the nodes of the two given subsets $U, V$ of the bipartite graph onto each other using the maximum weight matching function $mwm$ (see Definition \ref{eqn:maximum_weight_matching_function}), and compute the weighted similarity of the given list of the traces by $sim_{list}:\Phi^*\times\Phi^*\rightarrow\mathbb{R}$. For $\phi,\phi'\in\Phi^*$ it holds:
    \[
    sim_{list}(\phi,\phi') = \frac{1}{|\phi|} mwm(G^{\phi}_{\phi'}, w_{tax})
    \]
The two similarities are combined into a global similarity measure. To achieve this, two hyperparameters, $\alpha_1, \alpha_2 \in [0, 1]$ and $\alpha_1 + \alpha_2$, are introduced to weight the respective similarities:
\[sim_{trace}(\sigma,\sigma')=\alpha_1 *sim_{list}(\sigma_\phi,\sigma'_\phi)+ \alpha_2 *sim_{cf}(\sigma,\sigma')\]
This approach allows for flexible weighting of the trace and list similarities, enabling the method to be tailored to specific use cases.

\subsection{TS4NAP approach}
\label{sec:TS4NAP}
For next activity prediction, the TS4NAP function applies the previously described $sim_{trace}$ function to all traces in the event log $\mathcal{L}$, given an input trace $\sigma$. This produces a list of similarity values, where each trace in the log is assigned a value representing its similarity to the input trace. The list of traces, $\{\sigma_1', \sigma_2', \dots, \sigma_n'\}$, is then sorted by similarity. Afterward, it is filtered to retain only the $n$ most similar traces. 
For each trace $\sigma'$, the event $e \in \Sigma$ at position $|\sigma| + 1$ is extracted, and the list of unique events $[e_1, \dots, e_n]$, ordered by similarity, is returned. Finally, these $n$ most similar activities are used to predict the next activity corresponding to the input trace. 

From the retained set of most similar traces, the immediate next event of each trace is taken and identical events are grouped. Each candidate event receives a score equal to the sum of the similarity values of the traces that propose it. Candidates are then ranked by this score in descending order, and the top n events are returned as the set from which the prediction is made. If fewer than n unique candidates exist, all available candidates are returned. The procedure is fully deterministic for a fixed event log and input trace. 

If multiple retrieved traces have exactly the same overall similarity, ties are resolved deterministically by preferring (i) the trace with higher diagnosis-list similarity, then (ii) the trace with higher control-flow similarity, and finally by prioritizing more recent traces. When different next events are proposed by traces that remain tied, they remain as separate candidates and are scored as described above. Based on this, TS4NAP is defined as $TS4NAP:\Sigma^*\times 2^\mathcal{C}\times\mathbb{N}\rightarrow\Sigma^n$.

\begin{algorithm}[H]
\caption{TS4NAP Algorithm}
\label{alg:TS4NAP}
\KwIn{Input trace $\sigma$, event log $\mathcal{L}$, parameter $n \in \mathbb{N}$}
\KwOut{List of events $E$}

Define $\mathcal{L}' = \left\{ \sigma^\prime \in \mathcal{L} \ \big| \ \left| \sigma^\prime \right| \geq \left| \sigma \right| + 1 \right\}$\;

Initialize $S \gets \emptyset$\;
\ForEach{$\sigma^\prime \in \mathcal{L}^\prime$}{
    Compute $s\left( \sigma^\prime \right) = sim_{\text{trace}}\left( \sigma, \sigma^\prime \right)$\;
    Add $\left( \sigma^\prime, \, s\left( \sigma^\prime \right) \right)$ to $S$\;
}

Sort $S$ in decreasing order based on $s\left( \sigma^\prime \right)$\;
Remove from $S$ each pair $(\sigma^\prime, s(\sigma^\prime))$ for which there exists a pair $(\sigma^{\prime\prime}, s(\sigma^{\prime\prime}))$ previously in the list such that $\sigma^{\prime\prime}(|\sigma| + 1) = \sigma^\prime(|\sigma| + 1)$\;

$S_n \gets \left[ \left( \sigma^\prime_1, \, s_1 \right), \left( \sigma^\prime_2, \, s_2 \right), \dots, \left( \sigma^\prime_n, \, s_n \right) \right]$\;

\For{$i \gets 1$ \KwTo $n$}{
    $e_i \gets \sigma^\prime_i\left( \left| \sigma \right| + 1 \right)$\;
}

$E \gets \left[ e_1, \, e_2, \, \dots, \, e_n \right]$\;

\Return $E$\;
\end{algorithm}

\section{Case Study}
\label{sec:case_study}
In this chapter, the proposed approach is evaluated using 36 medical real-life event logs. These event logs are derived from the MIMIC-IV database and are described in Section \ref{sec:constructed_event_log}. Furthermore, two hypotheses are formulated to assess the effectiveness of the TS4NAP approach. These hypotheses are derived from the challenges and distinguishing characteristics from Section \ref{sec:introduction}. The hypotheses are defined as follows.  
\begin{itemize}
    \item \textbf{H1:} The prediction of the next activity can be improved by the integration of taxonomic knowledge.
    \item \textbf{H2:} The improvement which is achieved by integrating taxonomic knowledge varies for different primary diagnosis.
\end{itemize}
To confirm the hypotheses, the TS4NAP approach was implemented in Python utilizing NetworkX\footnote{\url{https://networkx.org/documentation/stable/tutorial.html}} a library used for network and graph analysis and the PM4Py\footnote{\url{https://pm4py.fit.fraunhofer.de}} library, which is used for process mining applications. Furthermore, the simple{\_}icd{\_}10{\_}cm\footnote{\url{https://github.com/StefanoTrv/simple_icd_10_CM}} library is used to handle the diagnostic codes, which are encoded in the ICD-10-CM classification from 2021. For the medical procedures, the 2021 version of the ICD-10-PCS taxonomy is used for handling the procedure codes. To the best of our knowledge, no library was directly applicable to the ICD-10-PCS taxonomy. The necessary functionalities to calculate the IC and the LCS, were implemented by the authors using Python. 

\subsection{Construction of Event Logs} 
\label{sec:constructed_event_log}
The analysis presented in this study utilizes data extracted from the MIMIC-IV dataset, focusing specifically on patient diagnoses and medical procedures. This section outlines the  used data, the criteria for event log construction and the post-processing of the constructed event log. The event logs are based on the tables \textit{diagnoses\_icd} and \textit{procedures\_icd} from the MIMIC-IV dataset. These tables are utilized to extract the diagnoses of the patients and the sequences of events that correspond to patients' hospitalizations. The diagnosis is encoded by using the ICD-10-CM taxonomy, and the medical procedures are encoded by using the ICD-10-PCS taxonomy. Notably, the full codes are employed for both medical codes without truncation, ensuring a detailed representation of medical events.

For the diagnoses, this study focuses on the primary diagnosis and the nine most important secondary diagnoses according to each patient. The primary diagnosis is identified by using the \textit{seq\_num} attribute and setting it equal to $1$. The \textit{seq\_num} attribute assigns a priority to the diagnosis, indicating its relative importance during hospitalization. Similarly, the next nine most significant diagnosis are identified by dropping all diagnosis where \textit{seq\_num} is greater than 10. However, it is acknowledged, following Johnson et al., that the accuracy of diagnosis ranking may not be meticulously maintained by billing departments \cite{Johnson2023}. Given the absence of timestamps for diagnoses in the \textit{diagnoses\_icd} dataset, an initial timestamp is assigned to position the diagnosis event at the beginning of each patient trace. For the procedures, on the other hand, all existing activities are considered. Furthermore, the event log contains only patient hospitalizations where at least one medical procedure was performed. Subsequent events are organized chronologically, with procedures sharing identical timestamps, sorted according to their \textit{seq\_num}.

Furthermore, the \textit{Taxonomy Type} attribute is added to each event or trace attribute, which is used to assign either ICD-10-CM or ICD-10-PCS as values to the event or trace attributes. This is done to ensure that the approach determines the appropriate taxonomy for each event or trace attribute. Also, an artificial END event is introduced to indicate the end of the trace and thus the patient treatment. This event also has a timestamp that ensures it is the last executed event in the trace. Afterward, the event log is filtered such that only traces representing categories of primary diagnoses with 500 or more patient hospitalizations are retained. The category corresponds to the first three letters of the diagnosis, as explained in Section \ref{sec:medical_taxonomies}. The filtering ensures a focus on sufficiently represented diagnostic categories, which is mandatory to check if the proposed approach can work. This is intended to rule out the possibility that the approach is not really applicable due to a lack of representative cases. After the filtering 36, categories of diagnosis are included. Next, to ensure comparability and for analytical clarity, separate event logs are constructed for each of the 36 identified diagnosis. This approach allows for the analysis to be confined to patient groups with a higher degree of homogeneity in primary diagnostic categories. Furthermore, it also stops patients with different diagnostic categories to be compared to one and other. 

\begin{figure*}[htb]
\centering
  	{\includegraphics[width=.7\textwidth]{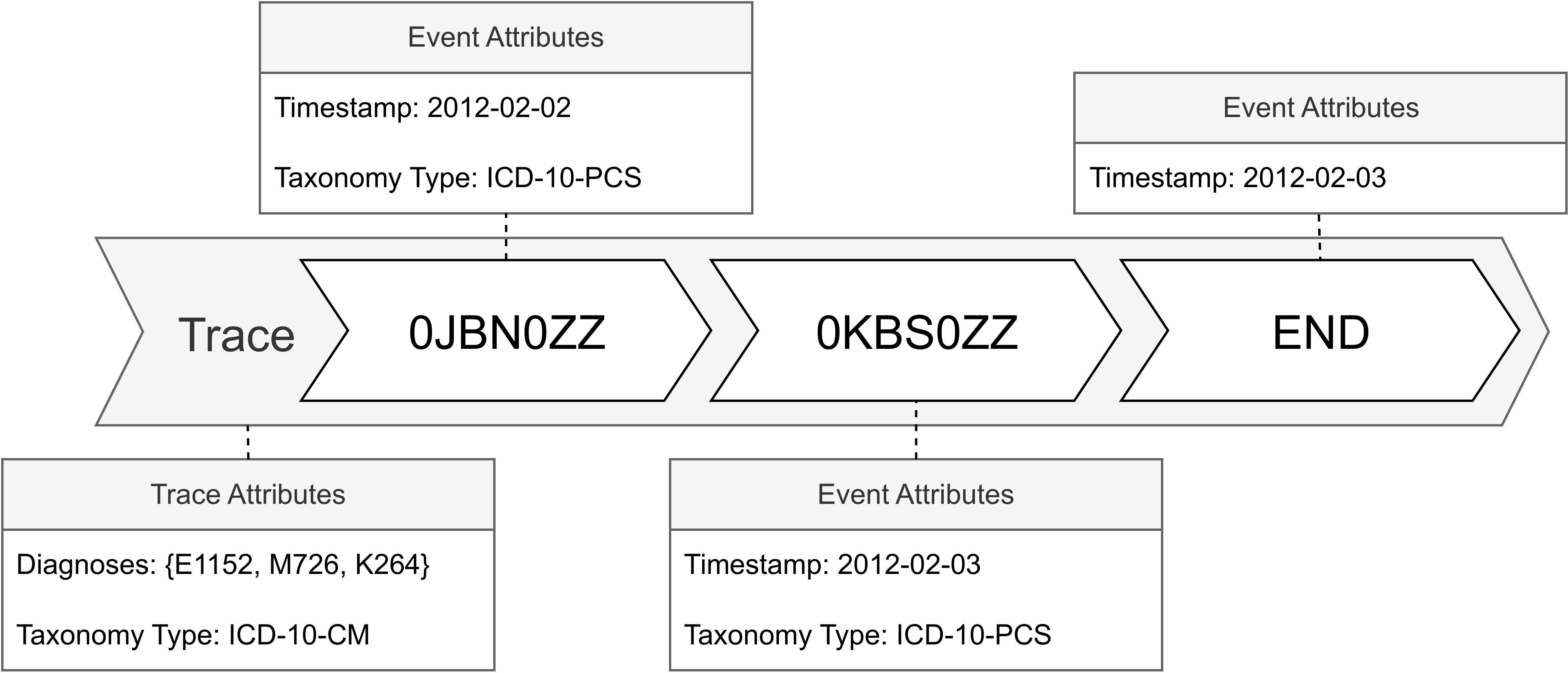}}
	\caption{Example trace from the event log, represented in graphical form\label{fig:trace_diagramm}}
\centering

\end{figure*}

Figure \ref{fig:trace_diagramm} shows a sample trace structure, where the event and trace attributes can be seen. The process shown in the middle of the Figure describe the control-flow, which denotes the activities as white arrows, which are performed in this trace. In the rectangular boxes, the event and trace attributes are described. It can be seen that the diagnoses attribute is coded as an ordered list of all diagnoses identified for a patient. Each of the 36 event logs contain many traces similar to the one shown in the figure but with varying length. 

\subsection{Statistical Analysis of Event Logs}
\begin{figure*}[htb]
\centering
  	{\includegraphics[width=0.8\textwidth]{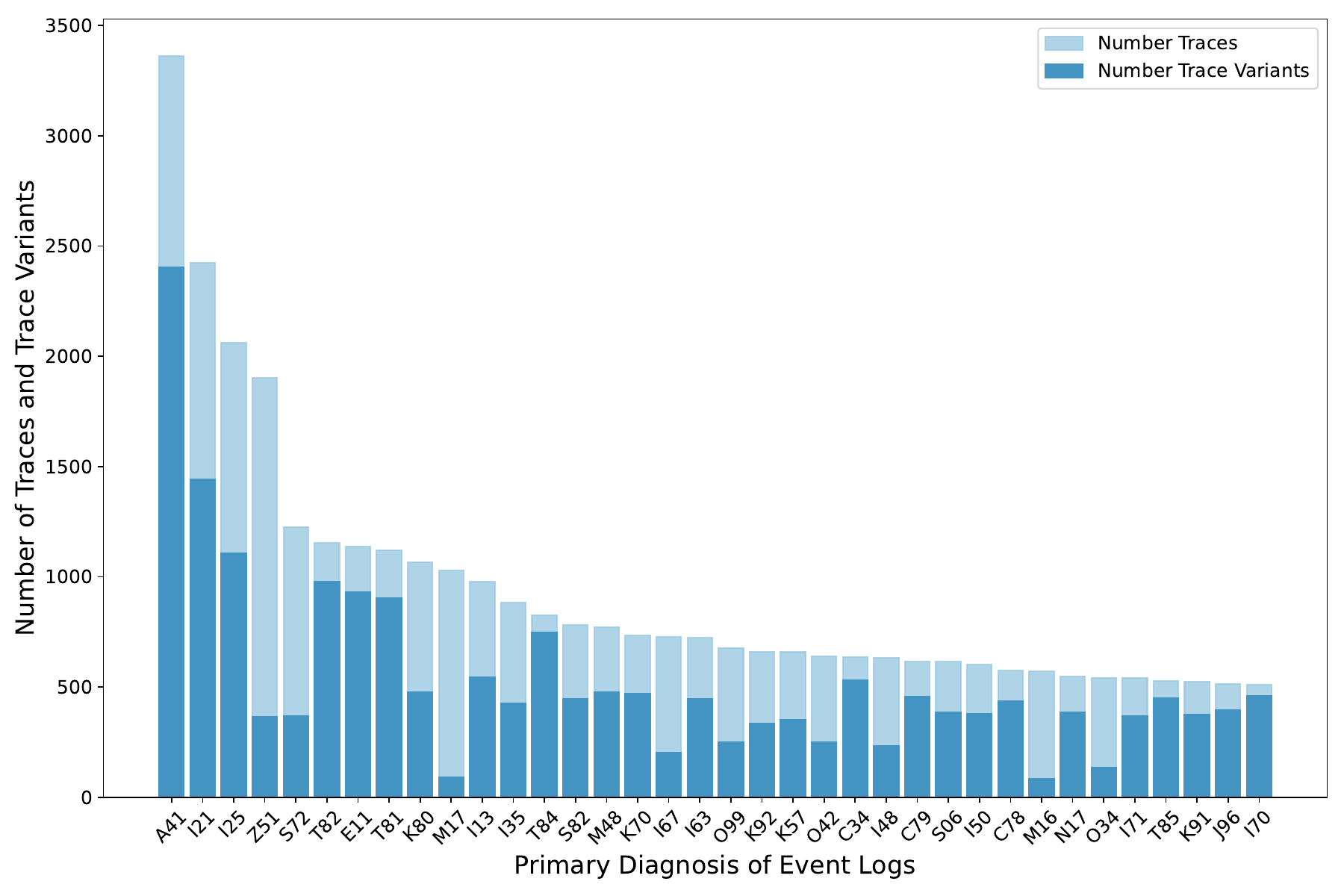}}
	\caption{Distribution of trace lengths and the number of trace variants shown for each event log on the same bar\label{fig:Trace_Variants}}
\centering

\end{figure*}

In this section, a descriptive analysis is conducted on the 36 event logs. Figure \ref{fig:Trace_Variants} highlights the number of trace lengths and the number of trace variants within each category of primary diagnosis for each event log. A trace variant can be understood as a trace that is unique in terms of the control-flow pattern, in the corresponding event log. The primary diagnoses \textit{A41} (Other sepsis) and \textit{I21} (Acute myocardial infarction) show a high number of traces. Where \textit{A41} has the most patients with 3361 and \textit{I70} (Atherosclerosis) has the least with 512. Also, \textit{I21} and \textit{I25} (Chronic ischemic heart disease) include 2424 and 2060 patient traces. Furthermore, the figure shows the number of trace variants within each primary diagnosis category. This is highlighted in the figure by the light gray bar in front of the darker bar that resembles the overall number of traces contained in the event log. It can be seen that for e.g., \textit{M17} (Osteoarthritis of knee) only has a little share of unique trace variants. On the contrary, \textit{I70} has a high number of unique traces in the event log. The same is true for \textit{T84} (Complications of internal orthopedic prosthetic devices, implants, and grafts) and \textit{T85} (Complications of other internal prosthetic devices, implants, and grafts). This could mean that patient treatments for these diagnoses are in general less standardized and harder to correctly predict. To put this into context, this could mean for \textit{M17} that it is more standardized or the process is less complex, and thus it can be easier predicted. 

Appendix \ref{fig:Mean_length_STD_app} indicates the average number of events per trace, categorized by the primary diagnosis of event logs. Some diagnoses, like \textit{I25} and \textit{K70} (Alcoholic liver disease), show longer trace lengths, suggesting more complex or long-term treatment patterns. On the other hand, conditions like \textit{M16} (Osteoarthritis of hip) and \textit{S72} (Fracture of femur) have shorter mean trace lengths, possibly indicating more straightforward or acute treatment scenarios. This variability in trace length can reflect the complexity of care, patient follow-up requirements, or the nature of the disease itself. Also, the black line indicates the variance of the mean trace length, indicating that for e.g., \textit{K70}, \textit{I71} (Aortic aneurysm and dissection) and \textit{S06} (Intracranial injury) show a high variability in treatment length and thus could be more complex to analyze. Contrary, \textit{M17} and \textit{M16} show low values of mean and standard deviation for trace length, which can indicate they are easier to analyze in a next event prediction setting.

Appendix \ref{fig:Number_Diagnosis_app} showcases the diversity of secondary diagnoses associated with primary diagnosis categories in patient traces. \textit{A41} and \textit{Z51} (Medical care involving chemotherapy, radiotherapy, and rehabilitation) stand out with the highest numbers of unique secondary diagnoses, suggesting these conditions often co-occur with a wide range of other health issues. This could indicate the complexity of managing these patients, who may have multiple comorbid conditions requiring simultaneous treatment. Thus, this can make it harder to predict the correct next activity if these diagnoses are present. On the contrary, \textit{O34} (Maternal care for abnormality of pelvic organs), \textit{O42} and \textit{O99} (Other maternal diseases classifiable elsewhere but complicating pregnancy, childbirth, and the puerperium) have the least amount of different secondary diagnoses.

Appendix \ref{fig:Number_Events_app} reflects the number of medical events, in the case of PCS codes, that patients with each primary diagnosis undergo. \textit{A41} and \textit{T81} (Complications of procedures, not elsewhere classified) have the highest numbers of unique events with 1742 and 1179 respectively, underscoring the complexity and variability of the medical interventions present for these primary diagnoses. On the contrary, \textit{O42} and \textit{M16} have the least number of unique events, with 97 and 88 performed medical procedures. The distribution of events according to the different primary diagnoses suggests that some primary diagnoses need a higher amount of different treatments according to the event logs used in this study. It can be assumed that it can be harder to predict the next activity if there are numerous unique events, in contrast to only a few events in an event log.

\subsection{Application of the TS4NAP Approach}
\begin{figure*}[htb]
\centering
  	{\includegraphics[width=0.8\textwidth]{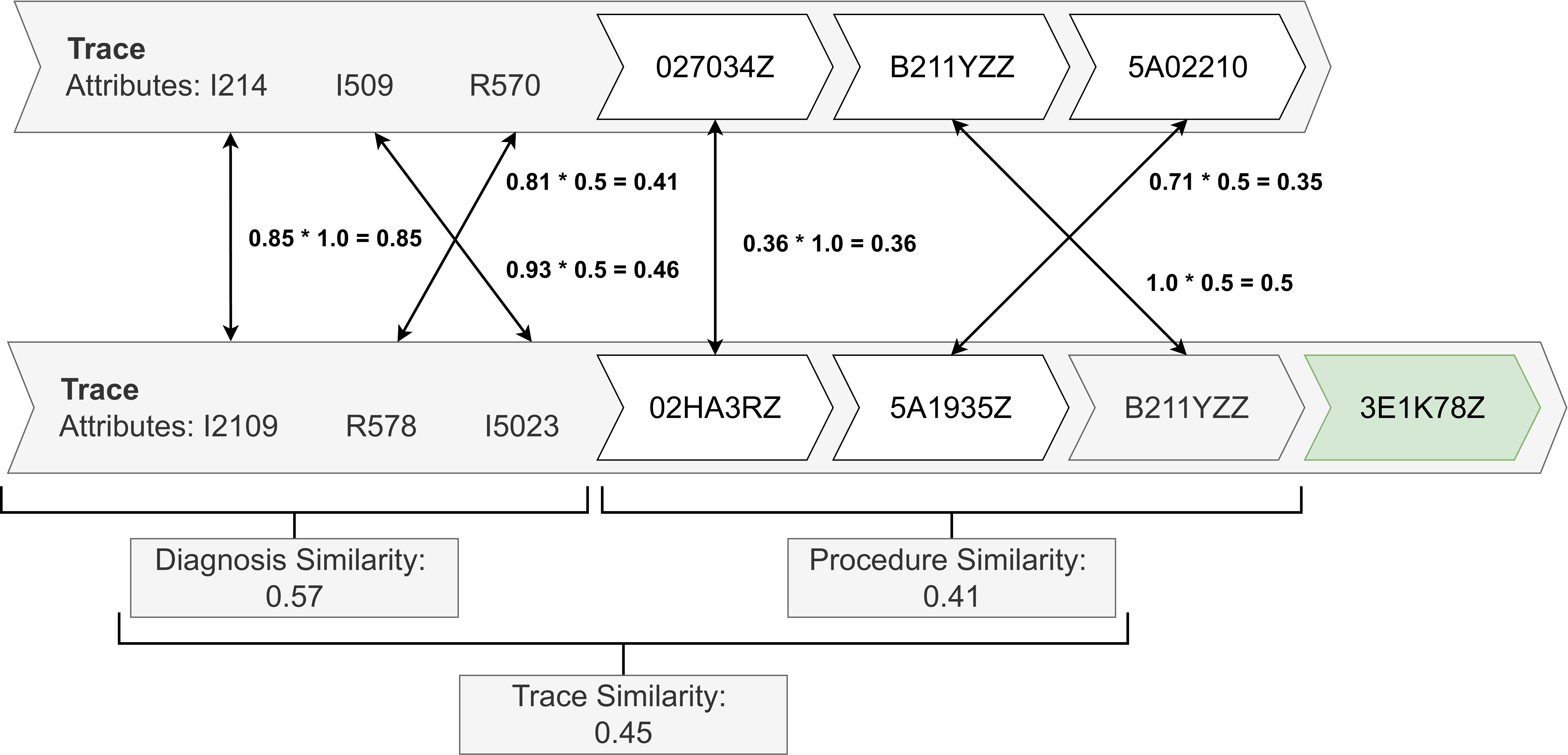}}
	\caption{The figure shows the calculation of similarity of two example traces. Beginning from the left side, the similarity between the list of diagnoses is calculated between the two traces. Moving to the right, the similarity between the medical procedures is calculated. The similarities are weighted according to the trace length and an aggregated similarity is calculated, which is depicted the as \textit{Trace Similarity}. The green arrow represents the prediction of the next activity, derived from the retrieved trace. \label{fig:example_calculation}}
\centering
\end{figure*}

In this section, it is explained how the TS4NAP approach is applied to the traces in the constructed event logs. As shown in Figure \ref{fig:trace_diagramm} each trace consists of a trace attributes, which contain the list of medical diagnoses and multiple events which resemble the medical procedures performed for the patient. To correctly assess the similarity of the traces contained in the event logs, the TS4NAP approach is applied in two steps and then aggregated to calculate a meaningful similarity. In the first step, the similarity for each list of diagnoses is calculated using $sim_{list}$. Therefore, the TS4NAP approach is applied only on the list of diagnosis (ICD-10-CM) and using the sequence number of the diagnoses as additional input for the order function. Thus, the different priorities of the primary diagnosis and the secondary diagnoses can be considered in the similarity calculation. As an example, for the diagnoses \textit{I214} and \textit{I2109} the function $sim_{\textsubscript{Sánchez}}(I214,I2109)$, which calculates the taxonomic similarity results in $0.85$ as shown in Figure \ref{fig:example_calculation}. For $w_{order}(I214,I2109)$ the result is, $1$ and the aggregated function therefore results in an edge weight of $0.85$. Thus, the importance of the different diagnoses as well as their similarity to each other is considered in the calculation. For the diagnoses \textit{R570} and \textit{R578} the taxonomic similarity results in $0.93$, but because there is a difference in the order of importance for the diagnoses the $w_{order}(R570,R578)$ results in $0.5$. Leading to an overall similarity for the diagnoses of $sim_{list} = 0.57$ In the second step, the $sim_{cf}$ function is applied for the control-flow, which is defined by the medical procedures (ICD-10-PCS). The similarity between the events is assessed, and the position of the medical procedure event in the trace is used as input for the order function. In this case the best allocation results in $sim_{cf} = 0.41$
Now, two values are calculated, the similarity between the list of diagnoses and the similarity between all the executed medical procedures. The $sim_{trace}$ function is used to aggregate the values calculated by 
$sim_{list}$ and $sim_{cf}$ using the hyperparameters $\alpha_1 and \alpha_2$ explained in Section \ref{sec:TS4NAP}, to assess the overall similarity of the trace. The hyperparameters will be adjusted based on the trace length. For $\alpha_1$, which sets the weight for $sim_{list}$, 
the weight is calculated by $\frac{1}{\text{trace length} + 1}$. On the other hand $\alpha_2$ is calculated by using the following formula, $\frac{\text{trace length}}{\text{trace length} + 1}$. For example, if there are three medical procedures and one list of diagnosis in the trace attributes, the value for $\alpha_1$ will result in $0.25$ and for $\alpha_2$ the value will result in $0.75$. This is only one method to calculate the hyperparameters, it is also possible to use static values, which do not change with varying trace length or choose parameters where the emphasis is higher for the list of diagnosis. In Figure \ref{fig:example_calculation} this process is clarified, and also it is shown how the retrieved trace is used for predicting the next activity. The edges between the events are representing the maximum weight matching, which is calculated. It is also visible which weight is associated with which edge, thereby considering the taxonomic similarity and the order function which represents the latter value in the calculation placed right of the edges in the figure. It can also be seen how the similarity for the diagnosis and for the medical procedures is aggregated, resulting in one value for the overall trace similarity. The green arrow in the figure shows the possible next prediction based on the retrieved trace.

\subsection{Evaluation}
For the evaluation, leave-one-out cross-validation (LOO-CV) is performed on each of the 36 event logs. In each fold, one trace is held out for testing and all remaining traces form the training set. For both training and testing, every trace is expanded into all proper, nonempty  prefixes, to capture the evolving event sequence. This enables next activity prediction at all positions within a trace rather than only at its end. For example, given the trace \([A,B,C]\), the prefix–target pairs  \([A]\rightarrow B\) and \([A,B]\rightarrow C\) are generated. The same procedure applies to all traces in each event log. 
The same procedure applies to all traces in each event log. Each individual prefix prediction is subsequently evaluated to assess the model's performance across different stages of the process. Two variants of our approach (TS4NAP) are compared: (i) \(\text{TS4NAP}_T\), which integrates taxonomic knowledge via the \(\text{sim}_{\text{Sánchez}}\) similarity (range \([0,1]\)), and (ii) \(\text{TS4NAP}_B\), which uses a boolean similarity \(\text{sim}_{\text{bool}}(a,b) = \mathbb{1}[a=b]\) and thus does not exploit taxonomic structure.

To measure the impact on the performance, while utilizing taxonomic knowledge for the calculation of event similarity, two different similarity functions are used and compared. Thus, two variants of the TS4NAP approach are defined. First, TS4NAP\textsubscript{T}, which integrates taxonomic knowledge by using $sim_{\textsubscript{Sánchez}}$. Second, TS4NAP\textsubscript{B}, which uses a boolean function for similarity calculation that determines whether two medical codes are identical by assigning a value of $1$ for identical codes and a value of $0$ otherwise. TS4NAP\textsubscript{B} presents the variant without incorporating taxonomic domain knowledge. 

To measure the overall accuracy of the TS4NAP approach, over all traces in the event log the \textit{Average Similarity} is introduced as an evaluation metric. The average similarity provides an aggregate metric that measures how good the next predicted activities align with the real activities for each trace in the event log, thereby considering the taxonomic relatedness of the next activities. Thus, the Average Similarity is defined as:  
\begin{itemize}
    \item $\frac{1}{|\mathcal{L}|} \sum_{j=1}^{|\mathcal{L}|} \underset{i \in \{1, 2, \ldots, n_j\}}{\max} (sim_{\textsubscript{Sánchez}} (R_j, P_{ji}))$, where

    \item $\mathcal{L}$ denotes the event log used as input,
    
    \item $R_j$ denotes the real next activity with the \(j\)-th trace in the event log and
    
    \item $P_{ji}$ denotes the next predicted activity, considering the \(j\)-th trace and \(i\)-th index of the set of predicted activities. $P$ can be seen as the set of predicted activities by the TS4NAP approach.

\end{itemize}
For both TS4NAP\textsubscript{T} and TS4NAP\textsubscript{B} the average similarity is selected as metric thus, guaranteeing the comparability of both methods considering the complete event logs. The choice of average similarity as a metric is motivated by the taxonomy-structured, fine-grained label space. Strict accuracy penalizes clinically acceptable near-misses as harshly as unrelated codes, whereas average similarity reflects graded similarity and was preferred by domain experts for practical relevance.

Moreover, a meaningful size of the predicted set of events was determined by conducting interviews with two domain experts. These experts were asked how many activities they would consider if a Clinical Decision Support System (CDSS) would recommend the activities. It was determined that the CDSS should ideally propose three to five  possible next activities. Furthermore, it is not desirable to predict only one possible next activity, as this would limit the medical staff in the selection of possible activities, which could lead to biased decisions if only one next possible event would be proposed. Thus, $n = 5$ was selected for this evaluation to address the heterogeneity in patient treatment.

\subsection{Results}
\label{sec:results}
This study evaluates the impact of integrating taxonomic knowledge on the average similarity of the next activity in treatment processes. It is hypothesized that (H1) the prediction of the next activity could be improved by the integration of taxonomic knowledge, and (H2) the extent of this improvement varies across different diagnoses. The evaluation was conducted using the 36 event logs constructed from the MIMIC-IV dataset.
\begin{figure*}[htb]
\centering
  	{\includegraphics[width=0.85\textwidth]{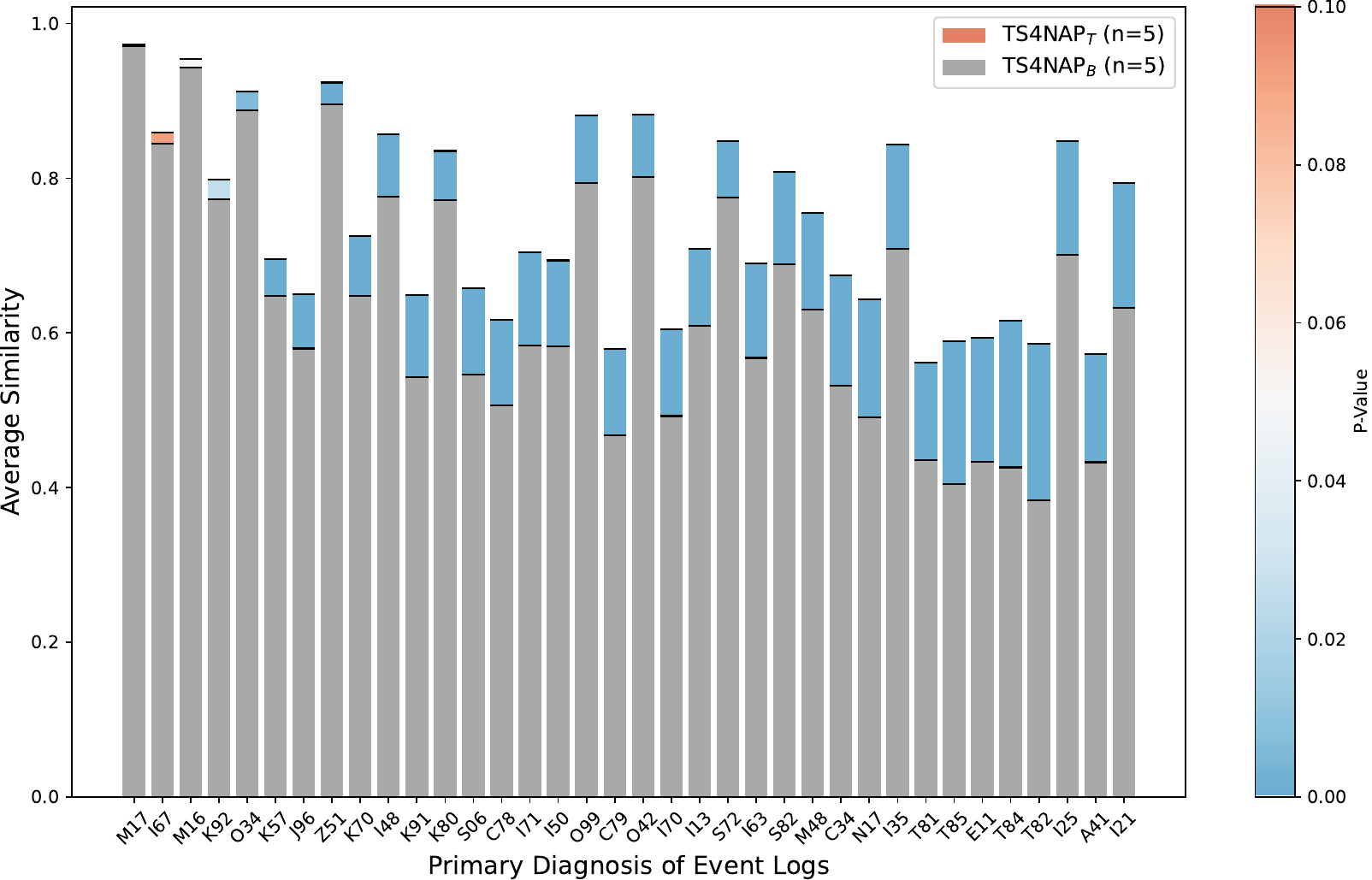}}
	\caption{Overlapping bar plot comparing the performance of the TS4NAP\textsubscript{T} approach to the baseline TS4NAP\textsubscript{B} approach across the constructed event logs. Gray bars represent the TS4NAP\textsubscript{B} approach average similarity scores, while colored bars for the TS4NAP\textsubscript{T} approach indicate the statistical significance of improvement as well as the similarity scores, with colors transitioning from red (less significant) to blue (more significant). \label{fig:overlapping_performance}}
\centering

\end{figure*}
The analysis revealed a significant improvement in the prediction of next activities with the integration of taxonomic knowledge using the TS4NAP approach. The integration significantly improved the average similarity scores for predictions of TS4NAP\textsubscript{T} compared to those made without taxonomic integration of TS4NAP\textsubscript{B}, which can be understood as baseline method. This enhancement was observed across the constructed event logs, which represent various primary diagnoses, showcasing the TS4NAP approach's effectiveness. Furthermore, the statistical significance of these results, validated through a one-sided t-test, are displayed in Figure \ref{fig:overlapping_performance}. In this figure, an overlapping bar plot is presented, where the baseline approach TS4NAP\textsubscript{B} is represented by gray bars. The bars representing the TS4NAP approach TS4NAP\textsubscript{T} are colored in a gradient from red to blue, indicating the significance of the p-values. The color spectrum from red to blue signifies the transition from less significant to more significant outcomes. The Figure, reveals that the TS4NAP approach yielded significantly better outcomes for 34 of the examined event logs. In only two instances (\textit{M17}, \textit{I67}) was the p-value above 0.05, indicating that in these specific scenarios, the TS4NAP method did not significantly outperform the baseline. In a considerable number of event logs, particularly for 15 specific primary diagnoses, average similarity scores exceeded 75\%. The highest recorded average similarity score reached 97\% (\textit{M17}), with the lowest at 56\% (\textit{T81}, Complications of procedures, not elsewhere classified), and an overall average of 74\%. A table with the detailed results can be seen in \ref{app:full_table}.

\begin{figure*}[htb]
\centering
  	{\includegraphics[width=0.85\textwidth]{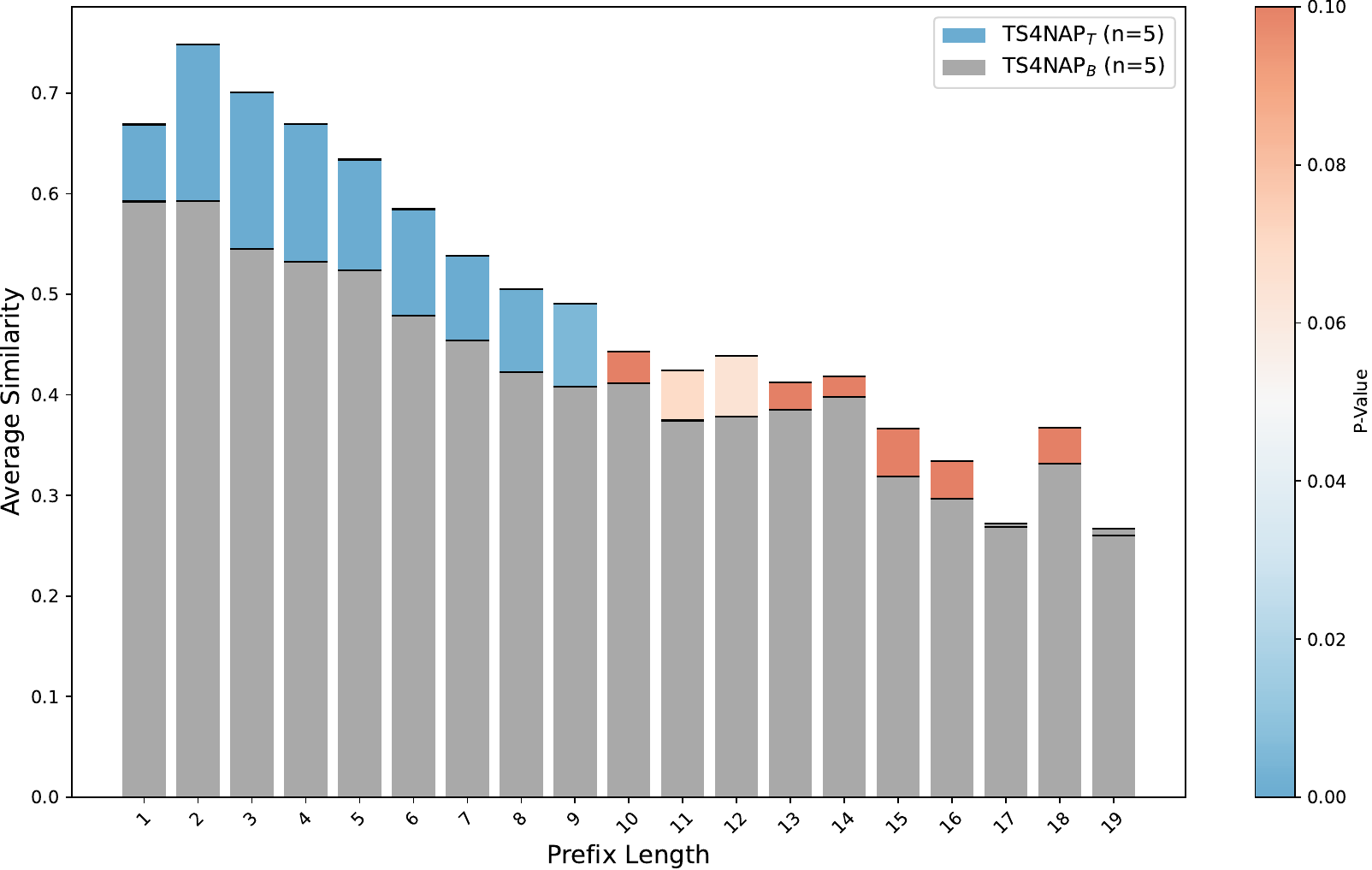}}
	\caption{Overlapping bar plot comparing the performance of the TS4NAP\textsubscript{T} approach to the baseline TS4NAP\textsubscript{B} approach across prefix lengths. Gray bars represent TS4NAP\textsubscript{B} average similarity scores, while colored bars for TS4NAP\textsubscript{T} indicate both the average similarity scores and the statistical significance of improvement, with colors transitioning from red (less significant) to blue (more significant). \label{fig:overlapping_performance_index}}
\end{figure*}

Performance as a function of prefix length was analyzed by grouping instances by prefix length. Figure~\ref{fig:overlapping_performance_index} summarizes how average similarity changes with increasing prefix length and visualizes the statistical significance of the differences between the taxonomy-aware TS4NAP\textsubscript{T} and the baseline TS4NAP\textsubscript{B}. A monotonic degradation of average similarity is observed for both variants as prefixes become longer, accompanied by a sharp decline in sample size. For short prefixes, TS4NAP\textsubscript{T} consistently and significantly outperforms TS4NAP\textsubscript{B}; for example, at prefix length \(1\) the average similarity increases from \(59\%\) (TS4NAP\textsubscript{B}) to \(67\%\), at prefix length \(2\) from \(59\%\) to \(75\%\), and at prefix length \(9\) from \(41\%\) to \(49\%\). Beyond prefix length \(\ge 10\), improvements diminish and are no longer statistically significant.

\begin{figure*}[htb]
\centering
  	{\includegraphics[width=0.8\textwidth]{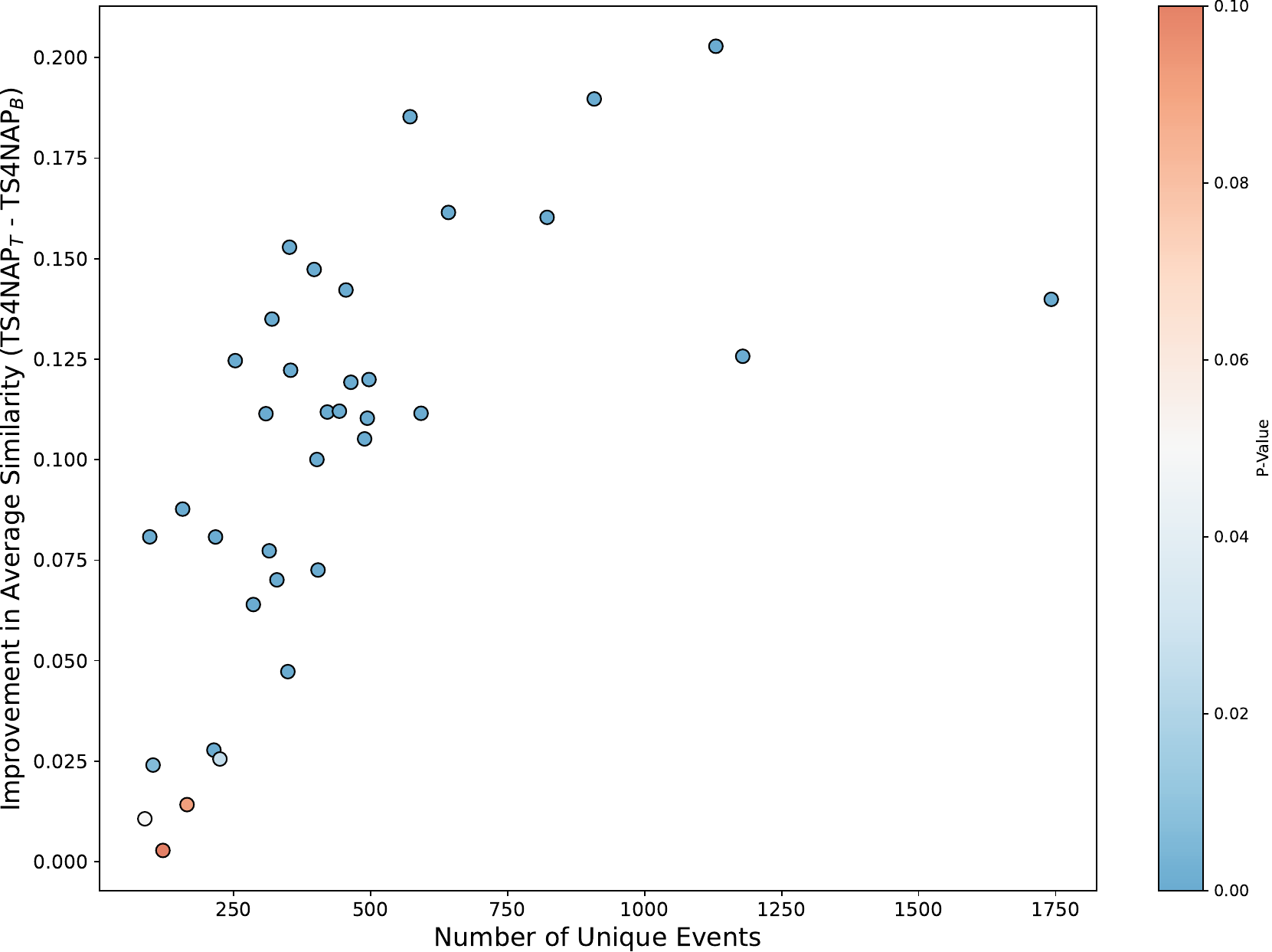}}
	\caption{Scatter plot that depicts the TS4NAP\textsubscript{T} approach's improvement over the baseline TS4NAP\textsubscript{B} approach in the context of unique events present in the constructed event logs. The dots represent the different event logs, plotted by the number of unique events (x-axis) and the improvement in prediction accuracy (y-axis), with color indicating the statistical significance from red (lower) to blue (higher).\label{fig:scatter_unique_events}}
\centering

\end{figure*}
Further examinations of the data revealed that the degree of improvement from integrating taxonomic knowledge into the predictive process varied across different diagnoses and the number of unique events within the specific event logs. Thus, Figure \ref{fig:scatter_unique_events} presents a scatter plot where each dot represents an event log. The dots are color-coded from red to blue, with the color intensity reflecting the significance of improvement represented by the p-value, when using the TS4NAP approach TS4NAP\textsubscript{T} over the baseline TS4NAP\textsubscript{B}. On this plot, the y-axis measures the difference in average similarity, while the x-axis accounts for the number of unique events in each event log. From this figure, it is observable that the event logs associated with a smaller number of unique events show no significant improvement when a taxonomy is integrated. Conversely, the benefit of using the TS4NAP\textsubscript{T} approach over TS4NAP\textsubscript{B} becomes more pronounced with an increase in the number of different events within the event logs. This suggests that the effectiveness of the TS4NAP approach is more significant in contexts with more unique events.

\begin{figure*}[htb]
\centering
  	{\includegraphics[width=0.8\textwidth]{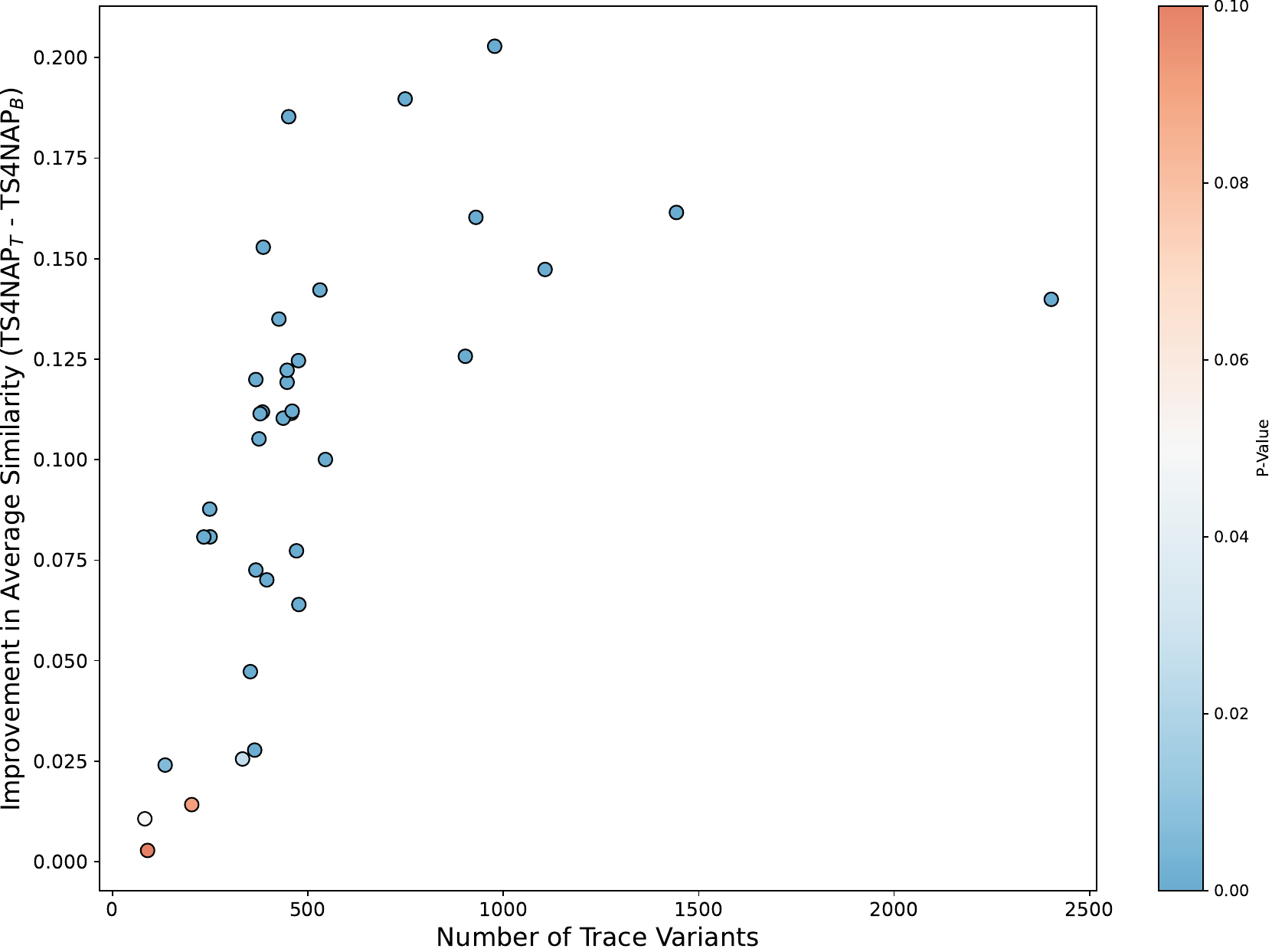}}
	\caption{Scatter plot that depicts the TS4NAP\textsubscript{T} approach's improvement over the baseline TS4NAP\textsubscript{B} approach in the context of trace variants present in the constructed event logs. The dots represent the different event logs, plotted by the number of unique events (x-axis) and the improvement in prediction accuracy (y-axis), with color indicating the statistical significance from red (lower) to blue (higher).\label{fig:scatter_trace_variants}}
\centering

\end{figure*}

Furthermore, the relationship between the trace variants and the significance of performance increase is analyzed. Figure \ref{fig:scatter_trace_variants} depicts a scatter plot that illustrates the differences between the TS4NAP\textsubscript{T} and the baseline TS4NAP\textsubscript{B} approach, plotted along the y-axis, against the number of trace variants on the x-axis. Trace variants refer to the number of unique event sequences of traces, considering only the sequence of events (control-flow) and not the attributes of each event. In this plot, the significance of the difference between TS4NAP\textsubscript{T} and TS4NAP\textsubscript{B} is indicated by the color coding of each dot, like in the previously discussed figures. Each dot represents one event log. Observations from Figure \ref{fig:scatter_trace_variants} indicate that event logs with a smaller number of trace variants tend to show lesser significant improvement when utilizing the TS4NAP\textsubscript{T} approach over TS4NAP\textsubscript{B}. However, as the number of trace variants increases, the TS4NAP\textsubscript{T} approach demonstrates a better performance compared to TS4NAP\textsubscript{B}, suggesting a positive dependency between the number of trace variants and the effectiveness of the TS4NAP\textsubscript{T} approach.
In summary, the analysis reveals that the TS4NAP\textsubscript{T} approach does not offer a significant advantage over TS4NAP\textsubscript{B} for event logs characterized by lower numbers of unique events or trace variants. Conversely, in scenarios with a higher number of unique events or a greater number of trace variants, the TS4NAP\textsubscript{T} approach significantly outperforms the baseline, indicating its effectiveness in more complex scenarios.

\section{Discussion}
\label{sec:discussion}
Using the TS4NAP approach to predict the next possible medical event is promising. In section \ref{sec:results} it  was shown that an improvement of the prediction could be achieved by the integration of medical knowledge held in taxonomies. Moreover, it was shown that the hypothesis H1 and H2 can be satisfied and thus confirm that the incorporation of medical knowledge present in taxonomies can enhance the predictive power for a next activity prediction in this case. Also, it was shown in H2 that for different primary diagnoses (each event log representing one) different values of performance are achieved. Additionally, the characteristic (D1) is addressed by utilizing event logs with a high diversity of events, diagnoses, and trace variants which can be seen in Section \ref{sec:constructed_event_log} and \ref{sec:results}. Moreover, (D8) is dealt with by using taxonomies enhanced with bipartite graph matching to make the predictions of the proposed approach more understandable and transparent. Furthermore, the predictions are explainable because the most similar patients are used for making the prediction. The challenge (C4) is addressed by evaluating it using real data in the form of the MIMIC-IV dataset.

A key observation from this study is the improvement in prediction accuracy across the primary diagnoses when employing the TS4NAP approach. This enhancement is significant in 34 of the 36 examined event logs. And an average similarity score of 74\% was observed considering all 36 event logs. This underscores the robustness of the TS4NAP approach in leveraging taxonomic knowledge, particularly in the domain of medical treatment planning. However, our analysis also revealed that the degree of this improvement is not uniform across all types of analyzed event logs. Specifically, the effectiveness of the TS4NAP approach appears to depend on the number of unique events and trace variants within the event logs. Event logs with a high number of unique events or trace variants exhibited more significant improvements in prediction accuracy. In addition, performance declines with increasing prefix length due to data sparsity and greater behavioral dispersion. TS4NAP\textsubscript{T} shows its most reliable gains for early to mid prefixes (prefix length \(\le 9\)), while improvements for very long prefixes are small and statistically insignificant. This suggests that the TS4NAP approach is suitable in complex medical scenarios with a high variety in treatments and diagnoses. A plausible explanation is that TS4NAP exploits semantic neighborhoods rather than relying on exact code matches. Taxonomy-based similarity treats semantically related codes as informative input for heterogeneous logs, mitigating sparsity that limits exact-match baselines. In addition, maximum-weight bipartite matching with order weighting accommodates local reorderings, so variation in control-flow has a smaller adverse effect on similarity. Finally, the diagnosis list functions as a semantic anchor, guiding retrieval toward clinically similar reference cases even when procedures diverge. These mechanisms are consistent with the empirical pattern that gains increase with the number of unique events and trace variants (Figures \ref{fig:overlapping_performance} and \ref{fig:scatter_unique_events}). 

TS4NAP offers recommendations, rather than treatment prescriptions, so predictions derived from coded histories may miss patient-specific contraindications, latent confounding, or rapidly evolving context, and can be sensitive to documentation quality, coding heterogeneity, and distribution shift; rankings are not calibrated for action thresholds, creating a risk of overreliance if used indiscriminately. At the same time, anticipating likely next steps can aid short-horizon scheduling, bed management, and timely allocation of equipment and staff, but operational use may inherit historical biases, overfit to local workflows, underperform when case mix shifts, or reduce flexibility if embedded too rigidly. Appropriate safeguards can include clinician oversight, local validation, routine performance monitoring, transparency about prediction confidence, rolling revalidation in operations, such that predictions function as a recommendation rather than binding directives.

The small improvement observed in event logs with fewer unique events or trace variants underscores a critical insight into the applicability of the TS4NAP approach. This suggests that the integration of taxonomic knowledge has lesser impact in a more straightforward context for, e.g., if a few unique events or trace variants are present in the event log. The additional computational complexity and resource investment required for integrating taxonomic knowledge into the TS4NAP approach may not always be justified, especially in less complex medical treatment scenarios where the capabilities of the TS4NAP method alone (only using TS4NAP\textsubscript{B}) yield satisfactory results. This understanding points to the importance of strategically applying taxonomic integration in scenarios where its benefits outweigh the higher computational demands, ensuring the optimization of resources and computational efficiency.

In order to further improve the TS4NAP approach, several challenges can be targeted in future research. While our findings highlight the potential of incorporating taxonomic knowledge into instance-based learning models, optimizing these models could lead to more practical usages. The TS4NAP approach, exhibits high computational complexity. This complexity impacts the runtime, particularly with large volumes of data. Thus, it would be beneficial to implement efficient retrieval mechanisms, that have a positive impact on the runtime, therefore techniques from Malburg et al. could be leveraged \cite{Malburg2021}. Such improvements could enhance the applicability of the approach in real-world scenarios. Moreover, for the $w_{order}()$ function a simple metric is introduced which respects the ordering of the events in a trace. It would be beneficial to implement other, more advanced techniques like Smith-Waterman algorithm to respect the ordering of events in a trace using alignments \cite{smith.1981}.  
Furthermore, integrating additional knowledge sources, including patient attributes such as gender, age and event-based attributes like timestamps, could further refine the predictive capabilities of the TS4NAP approach. The current data set was selected such that as many events and attributes as possible have a dependency on a taxonomy, in order to better validate the proposed approach. Expanding the dataset to include these multi-perspective data points may enhance the reliability of predictions.
The current implementation of the TS4NAP approach utilizes $sim_{\textsubscript{Sánchez}}$ for evaluating taxonomic similarity. Exploring alternative similarity measures that could be applied within the same taxonomic framework might reveal more optimal methods for enhancing next-activity prediction. 
The weights are designed to be deterministic and interpretable. A full sensitivity analysis across alternative order-penalty schemes, aggregation schedules, and taxonomic similarity measures lies beyond the scope of this study and will be addressed in future work.

Lastly, the integration of deep learning techniques, such as Long Short-Term Memory (LSTM) networks or sequence-to-sequence models, presents a promising approach for improving next activity prediction in complex scenarios. Enhancing the proposed approach with these models could increase the quality of predictions while still obtaining the explainability of those predictions. Therefore, the loss function of the deep learning models could be enhanced with taxonomic knowledge.
This study opens several directions for future work to improve the predictability of the proposed approach in real-life scenarios. By addressing these challenges, it is possible to develop a more accurate and robust approach for predicting the next activity in complex scenarios.

\section{Conclusion}
\label{sec:conclusion}
In this study, the TS4NAP approach is introduced, which leverages medical knowledge by utilizing the ICD-10-CM and ICD-10-PCS taxonomy, to improve the similarity-based next activity prediction and make it explainable for healthcare professionals. For the evaluation of this approach, 36 event logs were constructed from the MIMIC-IV database. The TS4NAP approach was applied to all 36 constructed event logs, and a detailed evaluation considering the complexity of the event logs as well as the performance of the TS4NAP approach was conducted, considering a variant of TS4NAP with medical knowledge and another variant without any additional coded knowledge. The results of the evaluation demonstrate the effectiveness of the proposed approach in improving the next activity prediction in complex environments. Furthermore, it was demonstrated that, the proposed approach creates meaningful predictions by leveraging taxonomies and past patient cases.  Future work can focus on extending the proposed approach by incorporating patient attributes and event-based features. Furthermore, to enhance the predictive quality of the TS4NAP approach, it can be investigated how deep learning approaches, such as LSTMs or sequence-to-sequence models can be integrated. 

\section*{Funding}
\noindent The research is funded by the German Federal Ministry of Research, Technology and Space (BMFTR) and NextGenerationEU (European Union) in the project KI-AIM under the funding code 16KISA115K.

\bibliographystyle{elsarticle-num} 
\bibliography{references}

\appendix

\onecolumn
\section{Additional Figures}
\label{app:additional_figures}
\begin{figure*}[htb]
\centering
  	{\includegraphics[width=0.65\textwidth]{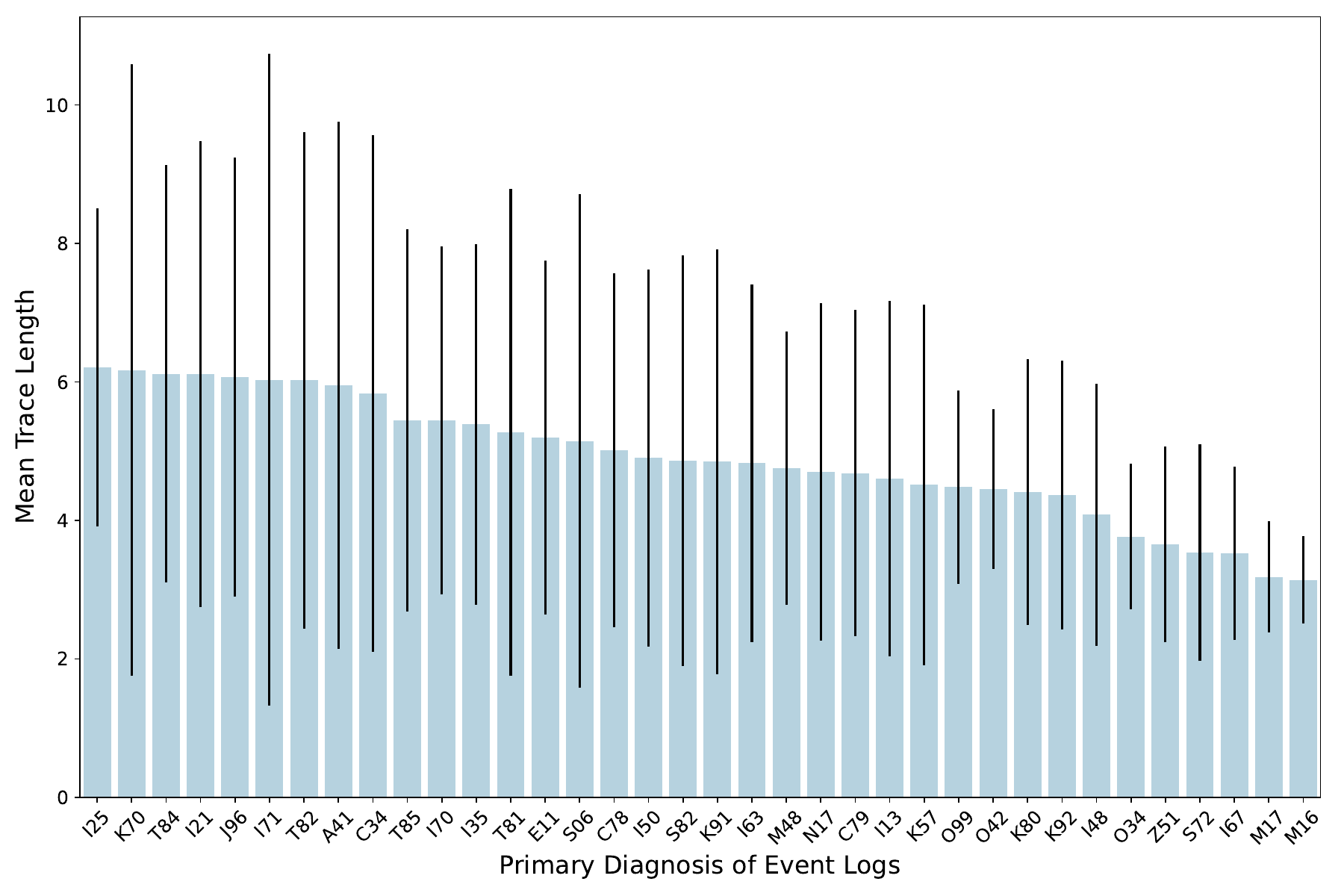}}
	\caption{Shows the mean trace length of the analyzed event logs, the line in the bar displays the standard deviation for the trace length.\label{fig:Mean_length_STD_app}}
\centering

\end{figure*}

\begin{figure*}[htb]
\centering
  	{\includegraphics[width=0.65\textwidth]{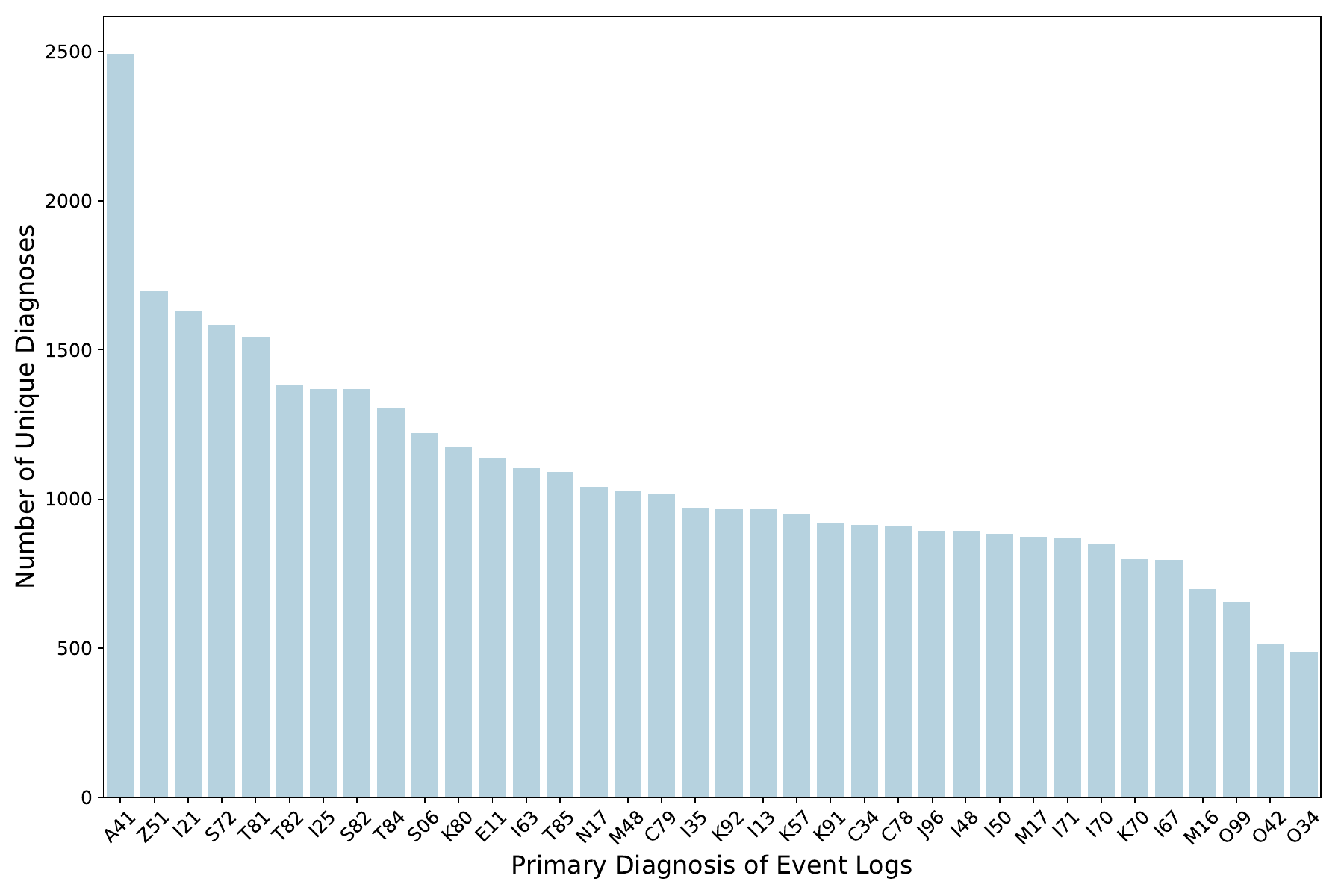}}
	\caption{Shows the number of unique secondary diagnoses for each of the 36 event logs. The X-axis describes the primary diagnoses of the event logs, and the Y-axis describes the number of unique secondary diagnoses related to the primary diagnoses.\label{fig:Number_Diagnosis_app}}
\centering
\end{figure*}

\begin{figure*}[htb]
\centering
  	{\includegraphics[width=0.7\textwidth]{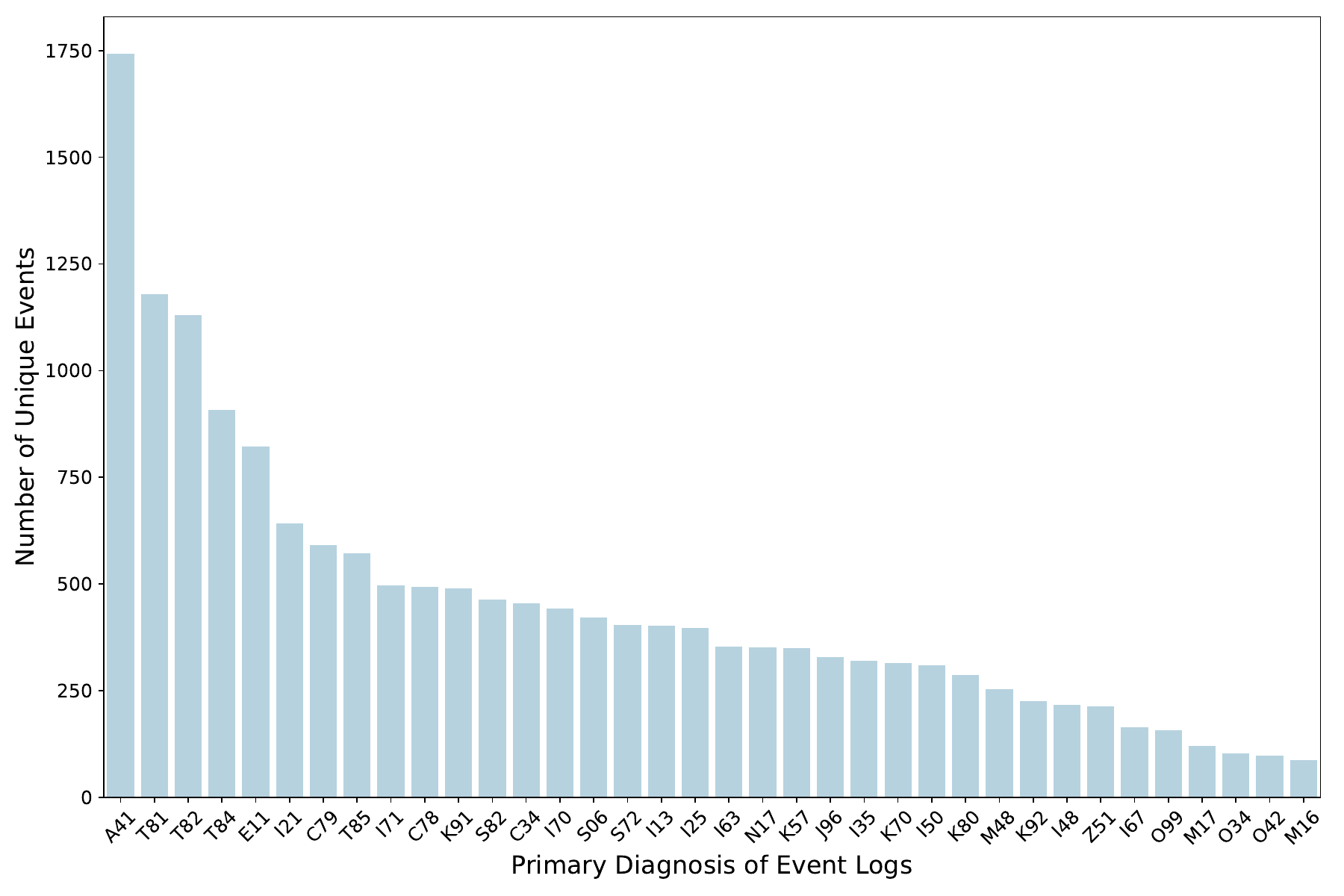}}
	\caption{Shows the number of unique PCS Codes encoded as events for each of the 36 event logs. The X-axis describes the primary diagnoses of the event logs, and the Y-axis describes the number of unique PCS-codes for the corresponding primary diagnoses. \label{fig:Number_Events_app}}
\centering
\end{figure*}

\onecolumn
\section{Detailed Results of the Evaluation}
\label{app:full_table}
\begin{landscape} 
\begin{table*}[!ht]
    \centering
    \tiny
    \begin{tabular}{|l|l|l|l|l|l|l|l|l|l|}
    \hline
        Prefix & \shortstack{Number \\ Traces} & \shortstack{Mean \\ Trace Length} & \shortstack{STD \\ Trace Length} & \shortstack{Average Similarity \\ TS4NAP\textsubscript{B} (n = 5)} & \shortstack{Average Similarity  \\ TS4NAP\textsubscript{T} (n = 5)} & \shortstack{Number \\ Trace \\ Variants} & \shortstack{Number \\ Unique \\ Events} & \shortstack{Number \\ Unique \\ Diagnosis} & P-Value \\ \hline
        A41 & 3361 & 5.948229694 & 3.807065485 & 0.432843806 & 0.572722416 & 2402 & 1742 & 2492 & 3.93124E-122 \\ \hline
        I21 & 2424 & 6.113448845 & 3.370002013 & 0.632663167 & 0.794152753 & 1443 & 642 & 1631 & 8.27112E-137 \\ \hline
        I25 & 2060 & 6.20776699 & 2.299679616 & 0.700619092 & 0.847923908 & 1107 & 397 & 1368 & 2.0673E-113 \\ \hline
        Z51 & 1902 & 3.651419558 & 1.413821301 & 0.896021461 & 0.923777742 & 364 & 214 & 1696 & 1.71017E-07 \\ \hline
        S72 & 1225 & 3.533877551 & 1.567345515 & 0.775153482 & 0.847702892 & 367 & 404 & 1584 & 1.14849E-19 \\ \hline
        T82 & 1155 & 6.022510823 & 3.587168389 & 0.383311592 & 0.586137086 & 978 & 1130 & 1385 & 1.68877E-95 \\ \hline
        E11 & 1136 & 5.19278169 & 2.556977292 & 0.433674595 & 0.593943872 & 930 & 822 & 1135 & 1.79159E-60 \\ \hline
        T81 & 1121 & 5.268510259 & 3.516311279 & 0.435473758 & 0.561187522 & 903 & 1179 & 1544 & 2.71697E-36 \\ \hline
        K80 & 1067 & 4.41049672 & 1.92207538 & 0.771390614 & 0.835355371 & 477 & 286 & 1176 & 1.83858E-13 \\ \hline
        M17 & 1029 & 3.184645287 & 0.800606164 & 0.970315557 & 0.973107609 & 90 & 121 & 873 & 0.268153679 \\ \hline
        I13 & 976 & 4.606557377 & 2.565034719 & 0.608995291 & 0.709031455 & 545 & 402 & 965 & 2.33336E-19 \\ \hline
        I35 & 882 & 5.387755102 & 2.604324474 & 0.708817588 & 0.843786358 & 426 & 320 & 968 & 1.30954E-33 \\ \hline
        T84 & 825 & 6.117575758 & 3.014849002 & 0.426211399 & 0.615939444 & 749 & 908 & 1307 & 6.04503E-68 \\ \hline
        S82 & 782 & 4.863171355 & 2.969404234 & 0.689027564 & 0.808274574 & 447 & 464 & 1368 & 9.40109E-26 \\ \hline
        M48 & 770 & 4.750649351 & 1.973203015 & 0.630687197 & 0.755315438 & 476 & 253 & 1026 & 5.56088E-26 \\ \hline
        K70 & 734 & 6.171662125 & 4.420029225 & 0.64794684 & 0.725270112 & 471 & 315 & 800 & 9.80831E-09 \\ \hline
        I67 & 727 & 3.522696011 & 1.251652085 & 0.844467238 & 0.858657314 & 203 & 165 & 795 & 0.091640241 \\ \hline
        I63 & 725 & 4.824827586 & 2.58094192 & 0.567718991 & 0.689962921 & 447 & 354 & 1104 & 2.03778E-22 \\ \hline
        O99 & 678 & 4.480825959 & 1.400174466 & 0.79385006 & 0.881551358 & 249 & 157 & 656 & 2.89563E-15 \\ \hline
        K92 & 659 & 4.362670713 & 1.940733457 & 0.772986218 & 0.798530717 & 333 & 225 & 967 & 0.025016479 \\ \hline
        K57 & 658 & 4.515197568 & 2.606147756 & 0.647775675 & 0.695046852 & 353 & 349 & 949 & 8.63866E-05 \\ \hline
        O42 & 639 & 4.455399061 & 1.15321472 & 0.801810415 & 0.882606321 & 250 & 97 & 513 & 1.08109E-16 \\ \hline
        C34 & 635 & 5.836220472 & 3.733421171 & 0.53210241 & 0.674302864 & 531 & 455 & 912 & 2.32374E-28 \\ \hline
        I48 & 632 & 4.080696203 & 1.897067757 & 0.775803218 & 0.856570215 & 234 & 217 & 894 & 3.90242E-10 \\ \hline
        C79 & 615 & 4.682926829 & 2.356352123 & 0.467858554 & 0.579388706 & 458 & 592 & 1015 & 4.09641E-16 \\ \hline
        S06 & 615 & 5.146341463 & 3.56805161 & 0.54585516 & 0.657699819 & 384 & 421 & 1220 & 3.43474E-14 \\ \hline
        I50 & 601 & 4.901830283 & 2.72311341 & 0.582370808 & 0.693800762 & 378 & 309 & 883 & 3.99071E-15 \\ \hline
        C78 & 576 & 5.010416667 & 2.557151098 & 0.506505769 & 0.616820461 & 437 & 494 & 909 & 1.96978E-14 \\ \hline
        M16 & 571 & 3.141856392 & 0.631775044 & 0.943071808 & 0.953740434 & 83 & 88 & 697 & 0.049787756 \\ \hline
        N17 & 549 & 4.703096539 & 2.43626971 & 0.490594101 & 0.64343658 & 386 & 352 & 1040 & 1.80214E-28 \\ \hline
        O34 & 541 & 3.767097967 & 1.051149586 & 0.88816255 & 0.912191868 & 135 & 103 & 488 & 0.006479079 \\ \hline
        I71 & 540 & 6.02962963 & 4.707300202 & 0.584092413 & 0.704017866 & 367 & 497 & 871 & 1.41207E-14 \\ \hline
        T85 & 527 & 5.445920304 & 2.762530639 & 0.404273281 & 0.589575057 & 451 & 572 & 1092 & 3.29819E-39 \\ \hline
        K91 & 524 & 4.849236641 & 3.066842876 & 0.543337744 & 0.648500982 & 375 & 489 & 921 & 1.03711E-12 \\ \hline
        J96 & 515 & 6.069902913 & 3.172560951 & 0.579918427 & 0.650016726 & 395 & 329 & 894 & 1.36597E-06 \\ \hline
        I70 & 512 & 5.4453125 & 2.516664129 & 0.492568716 & 0.604625084 & 460 & 443 & 847 & 1.81714E-17 \\ \hline
    \end{tabular}
    \caption{The table shows the detailed results of the evaluation for each of the 36 constructed event logs, for both the TS4NAP\textsubscript{B} and TS4NAP\textsubscript{T} approach.}
\end{table*}
\end{landscape}





\end{document}